%% file: main.tex
\documentclass{article}

\usepackage[main, final]{neurips_2026}
\usepackage[utf8]{inputenc} % allow utf-8 input
\usepackage[T1]{fontenc}    % use 8-bit T1 fonts
\usepackage{hyperref}       % hyperlinks
\usepackage{url}            % simple URL typesetting
\usepackage{booktabs}       % professional-quality tables
\usepackage{amsfonts}       % blackboard math symbols
\usepackage{nicefrac}       % compact symbols for 1/2, etc.
\usepackage{microtype}      % microtypography
\usepackage{xcolor}         % colors
\usepackage{amsmath}
\usepackage{hyperref}
\usepackage[capitalize]{cleveref}  
\usepackage{adjustbox}
\usepackage[table]{xcolor}
\usepackage{graphicx}
\usepackage{wrapfig}

\title{\texorpdfstring{\textsc{V-Lumen}}{V-Lumen}: Visual Lookup Memory for Embedding Scaling in Vision-Language Models}

\author{%
  Miso Choi\textsuperscript{1,*} \quad
  Daekeun Kim\textsuperscript{1,*} \quad
  Eunji Kim\textsuperscript{2} \quad
  Jungbeom Lee\textsuperscript{1,\textdagger} \\[2pt]
  \textsuperscript{1}Dept. of Computer Science and Engineering, Korea University \quad
  \textsuperscript{2}Amazon \\[2pt]
  \texttt{\{miso8070, housekdk, jbeomlee\}@korea.ac.kr} \quad
  \texttt{eunjkim@amazon.com} \\[4pt]
  \textsuperscript{*}Equal contribution. \quad
  \textsuperscript{\textdagger}Corresponding author.
}

\definecolor{baselinegray}{RGB}{247,247,247}
\definecolor{oursblue}{RGB}{235,247,255}

\begin{document}

\maketitle

\begingroup
\renewcommand{\thefootnote}{}
\footnotetext{Daekeun Kim holds concurrent appointments as a Ph.D. Student at Korea University and as an AWS Principal AI Solutions Architect. This paper describes work performed at Korea University and is not associated with AWS.}
\footnotetext{Eunji Kim's work on this paper was done prior to joining Amazon.}
\endgroup

\input{sections/0_abstract}
\input{sections/1_introduction}
\input{sections/3_method}
\input{sections/4_experiments}

\input{sections/5_analysis}
\input{sections/6_relworks}
\input{sections/7_conclusion}

\newpage

\begin{ack}
This work was supported by the NRF(RS-2026-25488668, RS-2026-25621049) and IITP under the Leading Generative AI Human Resources Development(IITP-2026-RS-2024-00397085) grant, the artificial intelligence star fellowship support program to nurture the best talents (IITP-2026-RS-2025-02304828) grant,  (RS-2022-II220043,Adaptive Personality for Intelligent Agents), ITRC(Information Technology Research Center) grant (IITP-2025-RS-2022-00156295), (No. RS-2026-25522672, Development of Unified Reasoning Technology Mimicking Human Cognition for Hierarchical Understanding and Unbounded Problem Solving), and IITP-ICT Creative Consilience Program grant (IITP-2026-RS-2020-II201819) funded by the Korea government(MSIT). This research was supported by the ANCHOR through the Seoul ANCHOR Center, funded by the Ministry of Education (MOE) and the Seoul Metropolitan Government. (2026-ANCHOR-01-003-09). This work was partly supported by IITP, AI Computing Support Project for R\&D grant funded by the Korea government(MSIT) (High-Performance Research AI Computing Infrastructure Support at the 2 PFLOPS Scale, RS-2026-25505492)
\end{ack}

% \bibliography{references}
% \bibliographystyle{neurips_2026}
{\small
\bibliographystyle{unsrt}
\bibliography{references}
}

%%%%%%%%%%%%%%%%%%%%%%%%%%%%%%%%%%%%%%%%%%%%%%%%%%%%%%%%%%%%
\newpage
\appendix
\input{sections/8_appendix}

%%%%%%%%%%%%%%%%%%%%%%%%%%%%%%%%%%%%%%%%%%%%%%%%%%%%%%%%%%%%

\newpage
\input{checklist.tex}

\end{document}

%% file: sections/0_abstract.tex
\begin{abstract}

Embedding-level scaling expands the capacity of language models through lookup-based memory rather than by increasing model depth, width, or dense computation. Extending this idea to vision-language models (VLMs) is challenging because visual features are continuous, spatially structured, and query-dependent, unlike discrete text tokens that naturally support $n$-gram lookup. We introduce \textsc{V-Lumen}, a visual lookup memory for VLMs. \textsc{V-Lumen} discretizes visual features with a frozen codebook, builds spatially aware memory keys through spatial aggregation-based hashing, and modulates them with text-conditioned relevance signals; all addressing is non-parametric, and the retrieved embeddings are fused into the backbone through a gated residual pathway. On five reasoning-intensive benchmarks, \textsc{V-Lumen} substantially improves a frozen Qwen3-VL-2B backbone and transfers zero-shot to three benchmarks whose images are disjoint from the training data. At a matched parameter budget, moving part of a LoRA or MoE adapter's capacity into \textsc{V-Lumen} preserves accuracy on two backbones, indicating that external visual memory is an independent capacity axis rather than a stronger adaptation method. Because lookup cost is constant in table size and incurred only once at prefill, this capacity grows in storage rather than in per-token compute. Code is available at \url{https://github.com/miso-choi/V-Lumen}.
 
\end{abstract}

%% file: sections/1_introduction.tex
\section{Introduction}
Scaling vision--language models (VLMs) by widening or deepening the backbone pays a dense compute price at every forward pass. A natural question follows: \emph{can we grow a VLM's visual representation capacity without growing its per-token compute?}

Recent work on LLMs answers an analogous question with \emph{external memory lookup}. Engram~\cite{cheng2026conditional}, STEM~\cite{sadhukhan2026stem}, and SCONE~\cite{yu2025scaling} augment a language model with a large table of learned embeddings and address it through a non-parametric hash over token IDs, so capacity grows with the table size while per-token activated compute stays flat. These methods assume a discrete token alphabet. Visual features in a VLM are continuous, laid out on a 2-D patch grid, and their relevance depends on the question being asked, so a direct port of text-side lookup is not available.

We introduce \textsc{V-Lumen}, an external visual memory for a VLM. \textsc{V-Lumen} is built from three components, each of which resolves one obstacle above. First, a frozen codebook~\cite{jin2024lavit} discretizes continuous patch features into visual code IDs, giving the memory a stable, reusable address space. Second, \emph{Spatial Aggregation-based Hashing} (SAH) folds row-, column-, and region-level context into the integer hash key, so the same code at different locations reaches different memory slots. Third, \emph{Text-Conditioned Hashing} --- via \emph{Projected Relevance Hashing} (PRH) and its debiased variant \emph{Debiased Relevance Hashing} (DRH) --- modulates the key with a coarse text-to-visual relevance signal, so the same image patch can route to different memory slots under different questions. Retrieved memory cells are then fused into the backbone through \emph{Multi-Head Conditional Gating} (MHCG), which assigns an independent gate per cell. All addressing on the memory side is non-parametric; only the memory tables and the fusion layer carry gradients.

We evaluate \textsc{V-Lumen} on five reasoning-intensive vision-language benchmarks spanning chart, document, and multi-discipline understanding. Attached to a frozen Qwen3-VL-2B backbone, \textsc{V-Lumen} improves accuracy on all five benchmarks without updating any backbone weight, and it transfers zero-shot to three benchmarks whose images share no source with the training data. We then ask how this memory relates to adaptation methods that add capacity in other ways. At a matched total budget, combining a smaller LoRA or MoE adapter with \textsc{V-Lumen} performs on par with a larger adapter alone in Qwen3-VL-2B, and the MoE comparison carries over to a second backbone, InternVL3.5-2B. Part of an adapter's capacity can thus be moved into an external memory without loss in accuracy, and this memory has a property that adapter parameters lack: its rows are addressed by hashing and only a few rows are read per visual token, rather than being used in dense matrix multiplications for every token. The memory can therefore, in principle, reside in host (CPU) memory or SSD-backed storage and be fetched on demand, as in large embedding tables of industrial recommendation models~\cite{naumov2019dlrm,agarwal2023bagpipe,wilkening2021recssd} and text-side lookup memories~\cite{cheng2026conditional}, leaving GPU memory for the backbone and adapters. We therefore view \textsc{V-Lumen} as an \emph{independent capacity axis}: capacity that matches adaptation in accuracy while growing in storage rather than in per-token compute.

% \textcolor{magenta}{
% Our contributions are the following. We identify visual embedding scaling as a capacity axis for VLMs that is distinct from backbone widening, LoRA, and MoE, and show how to realize it with a frozen backbone. We give a concrete recipe --- frozen codebook, SAH, PRH/DRH, and MHCG fusion --- in which every address-side mechanism is non-parametric, so added capacity shows up as table rows rather than as per-token compute. We show the capacity axis is complementary to LoRA and MoE, matching or beating their parameter- and activation-matched variants. Finally, we show through logit-lens analysis that the mechanism of improvement is representational reorganization in the mid-stack, not just faster convergence to the same prediction.}
Our contributions are as follows. First, we extend lookup-based embedding scaling from discrete text tokens to continuous visual features, with a recipe in which every addressing mechanism is non-parametric: a frozen codebook for discrete addresses, spatial aggregation-based hashing for region-aware keys, and text-conditioned hashing for query-dependent keys. Second, we show that this visual memory forms an independent capacity axis. It improves a frozen backbone and transfers zero-shot to unseen image domains, and at the same total budget it matches LoRA and MoE adapters on two backbones while holding part of their capacity in an external memory rather than in adapter parameters. Third, we show that this capacity grows in storage rather than in per-token compute: lookup latency stays constant across a $40\times$ range of table sizes, and the memory adds a one-off prefill cost without changing per-token decode latency. Because the memory is addressed by hashing and read sparsely, it can in principle be served from host memory or storage, which we leave to future work.

%% file: sections/3_method.tex
\section{Method}
\label{sec:method}

% Overall workflow
% 1. Convert visual features into discrete codes via Codebook
% 2. Construct visual embedding memory (n_heads, d_heads)
% 3. Spatial aggregation for the memory key
% 4. Text Query-conditioned Retrieval
% 5. Conditional fusion of retrieved visual memory
% 6. LoRA Alignment

%-----------------------------------------------------------------------------
\label{sec:prelim}
% We briefly recall the two building blocks that \textsc{V-Lumen} reuses directly: an Engram-style lookup memory, and a frozen visual codebook.
We briefly recall the Engram-style lookup memory that motivates the design of \textsc{V-Lumen}.
\paragraph{Engram.}
% Engram is a conditional-memory module for language models. 
% For each $n$-gram order $n\!\in\!\mathcal{O}$ and each hash head $h\in\{1,\dots,H\}$, a deterministic memory key $\phi_{n,h}$ of the suffix $n$-gram token IDs is constructed via multi-head hashing, which retrieves one row of a learnable embedding from the lookup table $\mathbf{E}_{n,h}$. 
% The final vector for suffix $n$-grams is the concatenation of all $|\mathcal{O}|\cdot H$ retrieved embeddings across the tables, which is then conditionally fused into the backbone hidden state as residual.
% Engram scales the total memory capacity by enlarging the lookup tables, while keeping the number of retrieved entries per token fixed. 
% This realizes a conditional-memory axis of sparse capacity, complementary to dense parameter scaling and MoE-style conditional computation. 
% \textsc{V-Lumen} inherits this motivation in \S\ref{sec:method:memory} and extends it with visual discretization, spatial aggregation, and text-conditioned hashing.
Engram~\cite{cheng2026conditional} augments language models with deterministic $n$-gram lookup memory, where a fixed number of memory entries are retrieved per token and fused into the hidden state as a residual.
This scales accessible memory capacity without increasing the number of retrieved entries per token, providing a conditional-memory axis of sparse capacity, complementary to dense parameter scaling or MoE-style conditional computation.
% \textsc{V-Lumen} extends this lookup-based motivation to VLMs through visual discretization, spatial aggregation, and text-conditioned hashing.

\subsection{Overview}
\label{sec:method:overview}
\input{figures/main_figure}

% \textsc{[V-Lumen]} extends a frozen vision--language backbone with an external visual memory, adding a scaling axis that is \emph{complementary} to parameter adaptation: rather than tuning the backbone, we retrieve per-position embeddings from a memory whose addresses are non-parametric and whose contents are learned, and fuse them back into the backbone hidden states via a residual pathway. The design builds on Engram~\cite{cheng2026conditional} and specializes it to the vision side of a VLM through three components [FIG1-overview]: \emph{(i) Visual Embedding Scaling} (\S\ref{sec:method:memory}) discretizes continuous patch features into codebook indices and reads per-position memory vectors whose content is learned; \emph{(ii) Spatial Aggregation} (\S\ref{sec:method:spatial}) builds region-aware keys so lookup uses local neighborhoods rather than isolated patches; and \emph{(iii) Text-Conditioned Hashing} (\S\ref{sec:method:retrieval}) routes the same patch to different memory slots depending on the query. The retrieved memory is fused into the visual tokens through a gated residual (\S\ref{sec:method:fusion}), and an optional small LoRA adapts the frozen backbone to the injected residual.
\textsc{V-Lumen} expands VLM capacity through a sparsely read external visual memory, a capacity axis that is independent of adaptation-based methods such as low-rank adapters and MoE-style conditional computation.
We retrieve per-position (visual token) embeddings from a memory, and conditionally fuse them back into the backbone hidden states via a residual pathway. 
\textsc{V-Lumen} specializes it to the vision side of a VLM through three components, as shown in Fig.~\ref{fig:main_overview}: \emph{(i) Visual Embedding Scaling} (\S\ref{sec:method:memory}) discretizes continuous patch features into codebook indices and reads per-position memory vectors; \emph{(ii) Spatial Aggregation-based Hashing} (\S\ref{sec:method:spatial}) builds region-aware keys so that lookup uses local neighborhoods rather than isolated patches; and \emph{(iii) Text-Conditioned Hashing} (\S\ref{sec:method:retrieval}) routes the same patch to query-adaptive memory slots, so that different text queries over the same image read from different parts of the memory. The retrieved memory is fused into the hidden states through a multi-head conditional gating (\S\ref{sec:method:fusion}), and an optional small LoRA or MoE adapts the frozen backbone to the injected residual.

% Formally, let $f_{\theta}$ be the frozen backbone (Qwen3-VL-2B-Instruct in our experiments) with hidden dimension $D{=}2048$, and let $\mathcal{L}_{\text{inj}} = \{\ell_{1}, \dots, \ell_{L}\}$ be the set of LLM decoder layers at which \textsc{V-Lumen} is injected via \texttt{forward\_hook}s ($L{=}2$; layers $\{2, 15\}$). At each $\ell \in \mathcal{L}_{\text{inj}}$ the post-layer hidden state receives a residual update $\Delta \mathbf{h}_{\ell} \in \mathbb{R}^{D}$:
% \begin{equation}
%   \mathbf{h}_{\ell}' \;=\; \mathbf{h}_{\ell} + \Delta \mathbf{h}_{\ell},
%   \qquad
%   \Delta \mathbf{h}_{\ell} \;=\; \mathrm{Fuse}_{\ell}\!\bigl(\mathbf{h}_{\ell},\, \mathbf{e}_{\ell}\bigr),
% \end{equation}
% where $\mathbf{e}_{\ell}$ is the memory read from \S\ref{sec:method:memory}--\S\ref{sec:method:retrieval}. The backbone $f_{\theta}$ itself receives no new weights.
Formally, let $f_{\theta}$ be the backbone VLM with hidden dimension $D$, which is kept frozen unless otherwise specified.
Let $\mathcal{L}_{\text{inj}} = \{\ell_{1}, \dots, \ell_{L}\}$ denote the set of LLM decoder layer indices at which \textsc{V-Lumen} is injected, where $L$ is the number of fused layers.
We set $L{=}2$ in our main experiments and choose the injection layers according to the layer ablation in Appendix \ref{app:layer_ablation}. 
For each $\ell \in \mathcal{L}_{\text{inj}}$, the hidden state is updated by a residual memory contribution:
\begin{equation}
  \mathbf{h}_{\ell}' \;=\; \mathbf{h}_{\ell} + \Delta \mathbf{h}_{\ell},
  \qquad
  \Delta \mathbf{h}_{\ell} \;=\; \mathrm{Fuse}_{\ell}\!\bigl(\mathbf{h}_{\ell},\, \mathbf{e}_{\ell}\bigr),
\label{eq:fusion}
\end{equation}
where $\mathbf{e}_{\ell}$ denotes the memory representation retrieved by the mechanisms described in \S\ref{sec:method:memory}--\S\ref{sec:method:retrieval}.

%-----------------------------------------------------------------------------
\subsection{Visual Embedding Scaling with External Memory}
\label{sec:method:memory}

% A naive application of Engram to the vision domain is non-trivial. Engram constructs memory addresses from discrete $n$-gram patterns over text tokens, where each token is represented by a finite vocabulary index. Visual tokens, by contrast, are continuous vectors: directly using combinations of continuous visual vectors as memory keys would produce an extremely sparse and poorly reusable address space, because even visually similar regions would yield slightly different continuous vectors, making exact (or deterministic) memory lookup ineffective and memory storage prohibitively expensive.
A central challenge in visual embedding scaling is how to define reusable memory addresses for visual representations. 
Unlike language tokens, which are already represented as discrete vocabulary indices, visual tokens are continuous feature vectors. 
Directly using raw visual features as memory keys would lead to an overly sparse address space: even visually similar regions may produce slightly different features, making deterministic memory lookup unreliable and limiting memory reuse. 
Moreover, storing memories for such fine-grained continuous addresses would be prohibitively inefficient. 
We address this by introducing a discrete key space, where continuous visual features are mapped to compact code indices before memory lookup.

\paragraph{Discretization via a frozen codebook.}
We first discretize continuous visual features into a finite set of visual codes using a codebook $\mathbf{C} \in \mathbb{R}^{V \times d_{\text{cb}}}$, where $V$ is the visual vocabulary size and $d_{\text{cb}}$ is the codebook embedding dimension. We adopt the pretrained LaVIT codebook~\cite{jin2024lavit} ($V{=}16{,}384$, $d_{\text{cb}}{=}32$) and keep it frozen, using it as a fixed visual vocabulary. Given a patch hidden state $\mathbf{x} \in \mathbb{R}^{D}$ captured at layer $\ell_{1}$, we apply a frozen random projection $\mathbf{W}_{\text{proj}} \in \mathbb{R}^{d_{\text{cb}} \times D}$ with $\mathcal{N}(0, 1/D)$ entries, L2-normalize, and take the nearest-neighbor codebook index:
\begin{equation}
  \mathbf{z} \;=\; \frac{\mathbf{W}_{\text{proj}}\, \mathbf{x}}{\lVert \mathbf{W}_{\text{proj}}\, \mathbf{x} \rVert_{2}},
  \qquad
  s(\mathbf{x}) \;=\; \arg\max_{v \in [V]}\; \mathbf{z}^{\top} \bar{\mathbf{c}}_{v},
  \label{eq:quantizer}
\end{equation}
where $\bar{\mathbf{c}}_{v}$ is the L2-normalized $v$-th codebook row. 
Although $\mathbf{W}_{\text{proj}}$ is frozen and randomly initialized, Johnson--Lindenstrauss-style random projections approximately preserve pairwise distances with high probability, making them a standard choice for dimensionality reduction and nearest-neighbor indexing \citep{johnson1984jl,freksen2021jltransforms}. Thus, nearby visual features tend to receive similar code assignments, which is sufficient for our purpose --- the discretizer only needs to induce a stable and reusable partition of visual features, not reconstruct the original representation.
Applying \cref{eq:quantizer} to all $T$ visual patch hidden states $\{\mathbf{x}_{t}\}_{t=1}^{T}$ at layer $\ell_{1}$ yields the visual codebook ID stream $s^{\text{vis}} = \bigl(s(\mathbf{x}_{1}),\dots,s(\mathbf{x}_{T})\bigr) \in [V]^{T}$, where $T= H_{v}W_{v}$ is the number of visual tokens per image.

% Both $\mathbf{W}_{\text{proj}}$ and $\mathbf{C}$ are frozen and do not carry a gradient, so the visual token alphabet is identical in all ablations.

% Given the visual ID stream $s^{\text{vis}} \in [V]^{T}$ produced by \cref{eq:quantizer}, we first pass it through the spatial aggregation stage of \S\ref{sec:method:spatial} to obtain a per-position \emph{aggregated visual key} $\kappa^{\text{vis}}_{t}\in\mathbb{N}$ at each image position. The aggregated key is \emph{not} a codebook ID: it is the integer hash-key derived from the codebook ID at position $t$ by mixing in the axial/coord refinements of \S\ref{sec:method:spatial} (and, when enabled, the text-conditioned salt of \S\ref{sec:method:retrieval}). The aggregated key is what the Engram backbone~\cite{cheng2026conditional} subsequently hashes. Let $\mathcal{O} \subset \mathbb{N}$ be the set of orders the memory attends to (we use $\mathcal{O} = \{2, 3\}$), let $H$ denote the number of memory heads (we use $H{=}8$), and let $d_{\text{head}}$ denote the width of a single memory cell (we use $d_{\text{head}}{=}8$, so the total memory width is $d_{\text{mem}} = |\mathcal{O}| H d_{\text{head}} = 128$). For each order $n\in\mathcal{O}$ and each head $h\in\{1,\dots,H\}$, a fixed non-trainable index map $\phi_{n,h}$ turns the aggregated key $\kappa^{\text{vis}}_{t}$ into a row address of a learnable embedding table $\mathbf{E}^{\text{vis}}_{n,h}$, and the read per-position is the concatenation across all tables $|\mathcal{O}|\cdot H$,
Given this ID stream $s^{\text{vis}} \in [V]^{T}$, we construct a per-position visual memory key through the spatial aggregation and text-conditioned hashing stages described in \S\ref{sec:method:spatial} and \S\ref{sec:method:retrieval}. The codebook ID at each image position is first enriched with spatial information from axial and coordinate-based aggregation, and is then conditioned on the input text query to produce a per-position aggregated visual key (formally defined in \cref{eq:final-visual-key}). This key is not a codebook ID; rather, it is an integer hash key derived from the local visual code, spatial context, and text-conditioned retrieval signal.

\subsection{Spatial Aggregation-based Hashing (SAH)}
\label{sec:method:spatial}

% The Engram backbone of \S\ref{sec:method:memory} treats the codebook-ID sequence as a 1-D stream, discarding the 2-D structure of an image: semantically similar patches that land on different codes get different $n$-gram contexts and therefore different hash addresses. We mitigate this by building \emph{region-aware} keys before hashing. Let $\mathbf{G} \in [V]^{\,H_{v}\times W_{v}}$ denote the $H_{v}\!\times\!W_{v}$ grid of codebook IDs for a single image, in row-major order ($H_{v}{=}W_{v}{=}8$ in our experiments). We replace the identity ID stream fed to the Engram memory with an aggregated grid $\widetilde{\mathbf{G}}$ produced by two non-parametric refinements applied in sequence:
Visual memory keys should reflect not only which visual code appears at a patch, but also how that code is situated within the image.
% The same code can carry different meanings depending on its neighboring codes and absolute location, so a memory key based only on the per-patch code ID would be too context-free for region-level visual retrieval.
The same code can carry different meanings depending on its neighboring codes and absolute location; therefore, a memory key based only on the per-patch code ID lacks sufficient context for region-level visual retrieval.
Therefore, we construct \emph{spatially contextualized} visual keys that incorporate both local spatial context and absolute positional cues before memory lookup.
Let $\mathbf{G} \in [V]^{\,H_{v}\times W_{v}}$ denote the $H_{v}\!\times\!W_{v}$ grid of codebook IDs for a single image, reconstructed from the visual ID sequence in row-major order ($H_{v}{=}W_{v}{=}8$ for Qwen3-VL-2B).
Starting from $\mathbf{G}$, we produce an aggregated grid $\widetilde{\mathbf{G}}$ using two non-parametric refinements applied in sequence.

% \textbf{(a) Axial pooling.} Codebook IDs from LaVIT are not translation- or row/column-stable: two semantically similar patches on the same row may discretize to different IDs, which in turn produces different $n$-gram contexts and therefore different hash addresses. We counter this by injecting row/column summaries back into each cell. Let $\mathrm{mode}_{\text{row}}(i)$ and $\mathrm{mode}_{\text{col}}(j)$ denote the majority codebook ID along row $i$ and column $j$ of $\mathbf{G}$, and let $\beta \in [0,1]$ be a fixed mixing weight. We form an axial-mixed grid $\mathbf{G}^{\text{ax}}$ by combining the per-cell ID with the row/column modes and re-quantizing (rank-discretize back to an integer ID):

\textbf{Axial pooling.}
To capture coarse spatial context, we summarize visual codes along the horizontal and vertical axes and fold the summary back into each cell.
This allows each patch key to reflect row- and column-level context, which is useful for structured visual inputs such as charts, tables, and documents.
% We emphasize up front that codebook IDs are \emph{categorical}---arithmetic on raw IDs has no semantic meaning---so the axial stage is designed to produce a \emph{deterministic integer hash salt} per row and column, not a semantic average in codebook space.
% The resulting grid $\mathbf{G}^{\text{ax}}$ is subsequently consumed by the multi-order hash $\phi_{n,h}$ (\S\ref{sec:method:fusion}), which determines the final memory address.

Concretely, for a row/column index $u$ and a 1-D slice $\mathbf{s}\in[V]^{n}$ of codebook IDs (row: $\mathbf{s}=\mathbf{G}_{u,:}$, column: $\mathbf{s}=\mathbf{G}_{:,u}$), we define a summary
\begin{equation}
  \rho : [V]^{n} \to \mathbb{Z}_{\ge 0},
  \qquad
  \rho(\mathbf{s}) =
  \begin{cases}
    \mathrm{mode}(\mathbf{s}), & \text{if some codebook ID in $\mathbf{s}$ repeats,} \\
    \mathcal{H}\!\bigl(\mathrm{sort}(\mathbf{s})\bigr), & \text{otherwise,}
  \end{cases}
  \label{eq:axial-summary}
\end{equation}
where $\mathcal{H}(\cdot)$ is a deterministic, order-independent rolling hash of the sorted codebook IDs (details in \cref{app:refiners}).
We set $\rho_{\text{row}}(i) = \rho(\mathbf{G}_{i,:})$ and $\rho_{\text{col}}(j) = \rho(\mathbf{G}_{:,j})$.
The ``repeat'' condition means some codebook ID appears at least twice in the row (resp.\ column), so that $\mathrm{mode}(\cdot)$ is well-defined; when all $W_v$ (resp.\ $H_v$) codes in a row (resp.\ column) are distinct, the mode is ambiguous and $\mathcal{H}$ is used instead.
% This rolling-hash fallback ensures that rows/columns \emph{without} a dominant code still contribute a stable integer summary rather than an ill-defined mode.
% We note that $\rho(\mathbf{s})$ is \emph{not} itself a codebook ID: it coincides with a codebook index when the first branch fires, but is an out-of-vocabulary hash integer in the rolling-hash fallback branch.
% In either case, $\rho$ is consumed only as an input to the subsequent hash $\phi_{n,h}$ (\S\ref{sec:method:fusion}), so its value space need not align with $[V]$.
With a fixed mixing weight $\beta \in [0,1]$, we form the axial-mixed grid by
\begin{equation}
  \mathbf{G}^{\text{ax}}_{i,j} \;=\; \mathrm{round}\!\Bigl(
    (1-\beta)\,\mathbf{G}_{i,j} \;+\;
    \tfrac{\beta}{2}\bigl(\rho_{\text{row}}(i) + \rho_{\text{col}}(j)\bigr)
  \Bigr).
  \label{eq:axial}
\end{equation}
% We reiterate that the rounded integer on the left-hand side is \emph{not} interpreted as a new codebook ID; it is an integer hash key. 
Patches that share the same row summary $\rho_{\text{row}}(i)$ (or column summary $\rho_{\text{col}}(j)$) therefore receive a row/column-correlated offset in the hash key, so they are routed to systematically related memory addresses by $\phi_{n,h}$---without requiring codebook-space arithmetic to be meaningful.
This yields a useful invariance for horizontally or vertically coherent structures.
\cref{eq:axial,eq:axial-summary} introduce no trainable parameters; the only hyperparameter is $\beta$.

\textbf{Coordinate bias.}
While axial pooling incorporates row- and column-level context, it can also make visually similar codes less distinguishable across different absolute regions of the image.
We therefore inject coarse positional information into the integer key using a position-level salt.
Specifically, we partition the $H_{v}\!\times\!W_{v}$ grid into an $H_{p}\!\times\!W_{p}$ coarse position grid (we use $4{\times}4$ positions for the $8{\times}8$ grid) and let $p(i,j) \in \{0,\dots,H_{p}W_{p}-1\}$ denote the coarse position index of cell $(i,j)$.
With a fixed salt prime $\pi_{\text{crd}}$ (see \cref{app:refiners} for its value), we define
\begin{equation}
  \widetilde{\mathbf{G}}_{i,j} \;=\; \mathbf{G}^{\text{ax}}_{i,j} \;\oplus\; \bigl(p(i,j)\cdot \pi_{\text{crd}}\bigr),
  \label{eq:coord}
\end{equation}
where $\oplus$ denotes the XOR operation.
Coordinate bias is also non-parametric and operates purely on integer keys.
The resulting grid $\widetilde{\mathbf{G}}$ encodes the \emph{spatially aggregated} visual key, which is subsequently modulated by the text query in \S\ref{sec:method:retrieval} before being used as the final memory lookup address.

%-----------------------------------------------------------------------------
\subsection{Text-Conditioned Hashing (TCH)}
\label{sec:method:retrieval}

% The memory retrieval process described so far depends only on the image: the same visual code at the same spatial context retrieves the same memory representation regardless of the input question. However, the relevance of a visual region often depends on the text query --- the same image may require retrieving different visual evidence depending on whether the question asks about an object, a spatial relation, or a chart element. We therefore introduce \emph{Text-Conditioned Hashing} (TCH), which modulates the visual memory key with a coarse text-to-visual relevance signal so that different queries over the same image route to disjoint memory slots.

While image-only retrieval provides a fixed visual representation, the relevance of a specific region often depends on the input question. To account for this, we introduce Text-Conditioned Hashing (TCH), which makes the retrieval process query-aware. TCH modulates visual memory keys with text signals, allowing the same image region to point to different memory slots depending on the specific task or question.

TCH keeps the routing \emph{parameter-free}: it modifies the integer memory key using a deterministic hash operation. 
% This design follows the broader motivation of deterministic routing in sparse systems~\cite{roller2021hash,lewis2021base,zuo2022thor}, while applying it to query-conditioned visual memory indexing. 
Motivated by the deterministic routing in sparse systems ~\cite{roller2021hash,lewis2021base,zuo2022thor}, we apply it to query-conditioned visual memory indexing with two instantiations: \emph{Projected Relevance Hashing} (PRH), described below, and its debiased variant \emph{Debiased Relevance Hashing} (DRH); their exact forms and an ensemble are deferred to Appendix.~\ref{app:prh},~\ref{app:drh}, and Appendix.~\ref{app:ablation} ablates PRH, DRH, and their ensemble.

Let $\mathbf{h}_{\text{txt}} \in \mathbb{R}^{D}$ denote the input hidden state of layer $\ell_{1}$ at the last text token (i.e., the pre-injection representation), and let $\mathbf{h}_{\text{vis},k} \in \mathbb{R}^{D}$ denote the $k$-th image-patch hidden state at the same layer, where $k \in \{1,\dots,N\}$ and $N = H_v W_v$. Using two frozen random projections $\mathbf{W}^{Q}_{\text{prh}}, \mathbf{W}^{K}_{\text{prh}} \in \mathbb{R}^{P\times D}$, PRH computes a per-patch dot-product relevance score
\begin{equation}
  r_{k} \;=\; \frac{\bigl\langle \mathbf{W}^{Q}_{\text{prh}} \mathbf{h}_{\text{txt}},\, \mathbf{W}^{K}_{\text{prh}} \mathbf{h}_{\text{vis},k}\bigr\rangle}{\sqrt{P}}.
  \label{eq:tch-score}
\end{equation}
DRH is a debiased variant of PRH that directly computes text--visual dot-product scores without random projection and applies an EMA-based prior correction to reduce scoring bias; we provide the details in \cref{app:drh}.
We rank-bucket the scores into $b_k \in \{0,\dots,K{-}1\}$ with $K{=}8$, and reshape them into a text-conditioned bin grid $\mathbf{B} \in \{0,\dots,K{-}1\}^{H_v\times W_v}$. The bin grid is then folded into the spatial hash key via a salted XOR:
\begin{equation}
  \widetilde{\mathbf{G}}_{i,j} \;\leftarrow\; \widetilde{\mathbf{G}}_{i,j} \,\oplus\, (\mathbf{B}_{i,j}\cdot \pi_{\text{tch}}),
  \label{eq:tch-xor}
\end{equation}
with $\pi_{\text{tch}}$ a salt prime distinct from $\pi_{\text{crd}}$ introduced in \S\ref{sec:method:spatial} (see \cref{app:refiners} for both values). The same image patch now addresses different memory slots when the text query changes.

For each grid cell $(i,j)$, we set the final memory key as
\begin{equation}
    \kappa^{\text{vis}}_{i,j} = \widetilde{\mathbf{G}}_{i,j}.
    \label{eq:final-visual-key}
\end{equation}

\subsection{Conditional Memory Fusion}
\label{sec:method:fusion}

Given the final visual memory key $\kappa^{\text{vis}}_{i,j}$, 
\textsc{V-Lumen} retrieves a per-position memory vector from multi-order, multi-head lookup tables.
For each order $n\in\mathcal{O}$ and head $h\in\{1,\dots,H\}$, a deterministic hash map $\phi_{n,h}$ maps $\kappa^{\text{vis}}_{i,j}$ to a row address in the learnable visual memory table $\mathbf{E}^{\text{vis}}_{n,h}$.
For a visual position $t$ corresponding to grid cell $(i,j)$, the retrieved memory representation is
\begin{equation}
  \mathbf{e}^{\text{vis}}_{t}
  =
  \operatorname*{concat}_{n \in \mathcal{O},\, h \in \{1,\dots,H\}}
  \mathbf{E}^{\text{vis}}_{n,h}
  \!\left[\phi_{n,h}(\kappa^{\text{vis}}_{i,j})\right]
  \in \mathbb{R}^{d_{\text{mem}}}.
  \label{eq:visual-engram-read}
\end{equation}
This yields a fixed number of retrieved memory cells per visual token, while the total accessible memory capacity can grow with the table size.

The retrieved vector is then fused into the backbone hidden state through a gated residual pathway.
Rather than applying a single gate to the entire retrieved vector, we use cell-wise conditional gating, where each retrieved memory cell is weighted based on the current hidden state $\mathbf{h}_{\ell,t}$.
The resulting fused memory update is added as a residual as in \cref{eq:fusion}:
% \begin{equation}
%   \mathbf{h}'_{\ell,t}
%   =
%   \mathbf{h}_{\ell,t}
%   +
%   \Delta \mathbf{h}_{\ell,t},
%   \qquad
%   \Delta \mathbf{h}_{\ell,t}
%   =
%   \mathrm{Fuse}_{\ell}
%   \left(
%   \mathbf{h}_{\ell,t},
%   \mathbf{e}^{\text{vis}}_{t}
%   \right).
%   \label{eq:fusion-compact}
% \end{equation}
We provide the exact form of the multi-head conditional gating and the depthwise-convolution smoothing used in $\mathrm{Fuse}_{\ell}$ of \cref{eq:fusion} in \cref{app:fusion}.

\subsection{Retrieval Latency}
\label{sec:retrieval}
Because rows are addressed by hashing, the cost of a lookup should not depend on the size of the memory table. We test this by scaling the number of rows per cell from $0.2\times$ to $8\times$ of our default setting (81,920 rows per cell, used in all main experiments), a $40\times$ range, while keeping all other components fixed (Tab.~\ref{tab:retrieval_latency}). A lookup consists of two stages. \emph{Key construction} computes row indices from the spatially aggregated codes, coordinate bins, text-conditioned bins, and per-cell hashing, without accessing the table. \emph{Gather} then reads the memory embeddings at those indices. We measure both stages for one fusion layer with 64 visual tokens in bf16 on a single RTX 6000 Ada GPU, averaging 50 runs after 10 warm-up runs.

Across the $40\times$ range ($0.2\times$ to $8\times$), total lookup latency varies by less than 1\% (11.16--11.23\,ms), and gather time by at most 6\% (0.46--0.49\,ms). Most of the lookup time ($\sim$10.7\,ms) is spent on key construction, which never touches the table; the gather itself takes under 0.5\,ms. Lookup cost is therefore $O(1)$ in table size: enlarging the memory adds storage but not per-token computation, so the accessible capacity of \textsc{V-Lumen} is decoupled from its per-token compute.

\begin{table}[t]
\centering
\caption{Lookup latency versus memory table size for one fusion layer (64 visual tokens, bf16, RTX 6000 Ada; mean over 50 runs). Values in parentheses are relative to the default setting (1$\times$).}
\label{tab:retrieval_latency}
\small
\begin{tabular}{rrrcc}
\toprule
Rows / cell & Table params & Memory & Lookup total (ms) & Gather only (ms) \\
\midrule
% 16,384 {\scriptsize(0.2$\times$)}   & 8,433,472   & 16.9 MB  & 11.22{\scriptsize$\pm$0.15} {\scriptsize(1.000$\times$)} & 0.463 {\scriptsize(1.000$\times$)} \\
% 81,920 {\scriptsize(1$\times$)}   & 41,976,448  & 84.0 MB  & 11.16{\scriptsize$\pm$0.11} {\scriptsize(0.994$\times$)} & 0.471 {\scriptsize(1.016$\times$)} \\
% 327,680 {\scriptsize(4$\times$)} & 167,825,792 & 335.7 MB & 11.16{\scriptsize$\pm$0.11} {\scriptsize(0.995$\times$)} & 0.462 {\scriptsize(0.998$\times$)} \\
% 655,360 {\scriptsize(8$\times$)} & 335,605,248 & 671.2 MB & 11.23{\scriptsize$\pm$0.21} {\scriptsize(1.001$\times$)} & 0.491 {\scriptsize(1.059$\times$)} \\
16,384 {\scriptsize(0.2$\times$)}  & 8,433,472   & 16.9 MB  & 11.22{\scriptsize$\pm$0.15} {\scriptsize(1.005$\times$)} & 0.463 {\scriptsize(0.983$\times$)} \\
81,920 {\scriptsize(1$\times$)}    & 41,976,448  & 84.0 MB  & 11.16{\scriptsize$\pm$0.11} {\scriptsize(1.000$\times$)} & 0.471 {\scriptsize(1.000$\times$)} \\
327,680 {\scriptsize(4$\times$)}   & 167,825,792 & 335.7 MB & 11.16{\scriptsize$\pm$0.11} {\scriptsize(1.000$\times$)} & 0.462 {\scriptsize(0.981$\times$)} \\
655,360 {\scriptsize(8$\times$)}   & 335,605,248 & 671.2 MB & 11.23{\scriptsize$\pm$0.21} {\scriptsize(1.006$\times$)} & 0.491 {\scriptsize(1.042$\times$)} \\
\bottomrule
\end{tabular}
\end{table}

%% file: figures/main_figure.tex
\begin{figure}[t]
    \centering
    \includegraphics[width=\textwidth]{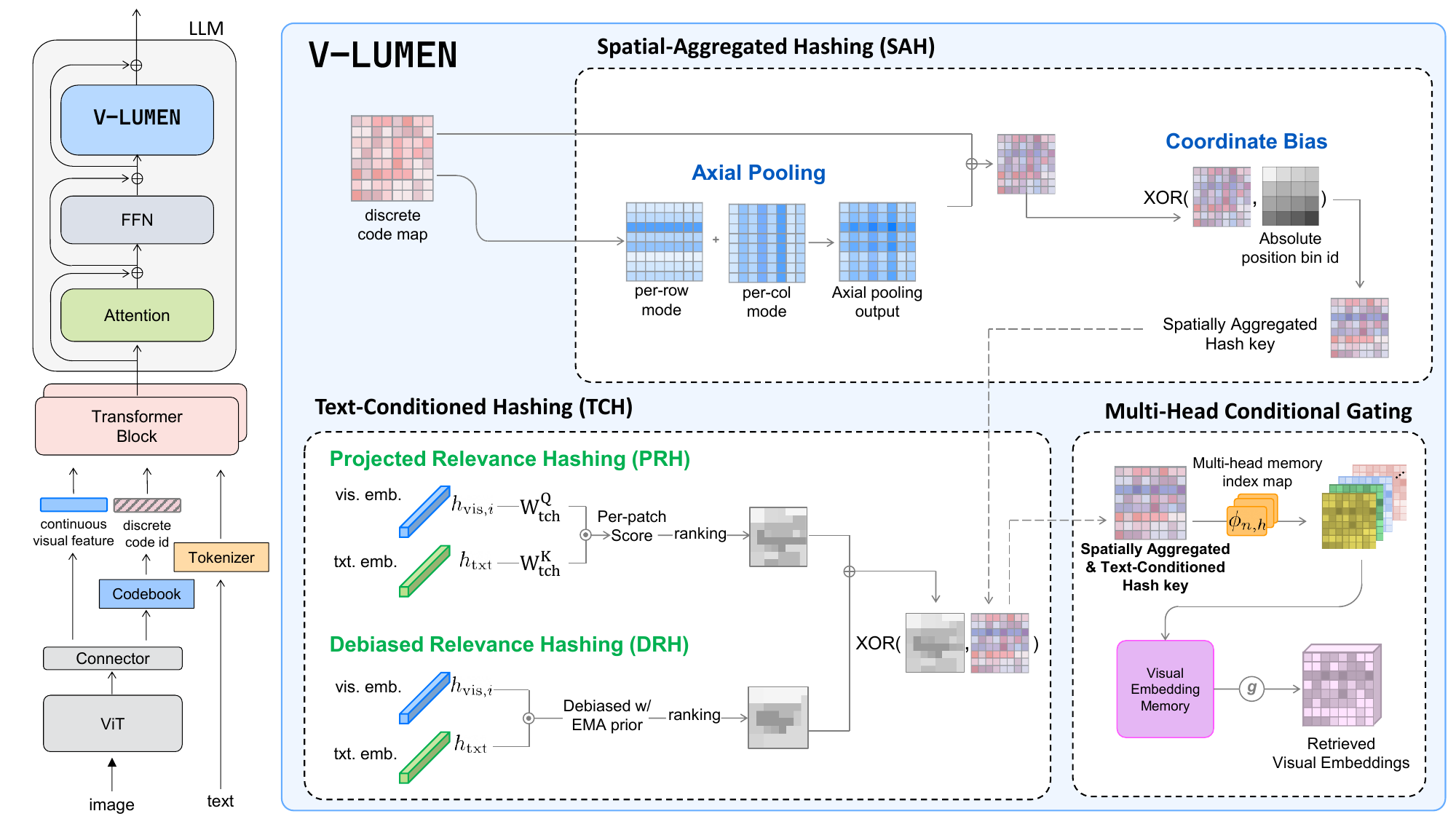}
    \caption{
    \textbf{Overview of \textsc{V-Lumen}.}
    \textsc{V-Lumen} retrieves visual memory embeddings using spatially aggregated and text-conditioned hash keys, then fuses them into the frozen VLM through gated residual updates.
    }
    \label{fig:main_overview}
\end{figure}

%% file: sections/4_experiments.tex
\section{Experiments}
\label{sec:experiments}
\subsection{Experimental Settings}
\paragraph{Training Datasets.}
% We construct the training data with the goal of encouraging the visual memory to capture reusable visual priors rather than memorizing task-specific answers. Since \textsc{V-Lumen} retrieves memory based on visual patterns, we organize the training data to cover diverse domains with visual pattern recurrence, including document layouts, tables, chart axes, OCR regions, object co-occurrences, and spatial relations.
The strategic design of our training data is paramount to ensuring that the visual memory captures reusable visual priors rather than merely memorizing task-specific answers. To facilitate this, we deliberately curate a training corpus from diverse domains characterized by high visual pattern recurrence, such as document layouts, tabular structures, and OCR regions. 
% By exposing the model to consistent structural motifs across various contexts, we incentivize the memory to internalize generalizable visual representations that can be effectively leveraged for downstream reasoning tasks.
This exposes the model to similar visual motifs across varied contexts, providing training signals that can make memory lookup beneficial for downstream reasoning tasks.

The training consists of two stages with different data mixtures: mid-training and SFT. The goal of mid-training is to form visual static priors in the memory, while the goal of SFT is reasoning alignment after memory retrieval.

% We use different data mixtures for Mid-training and SFT. The goal of mid-training is to form visual static priors in the memory. Accordingly, we organize the mid-training data into four groups: document-heavy data, spatial-structure data, diagram/chart/UI-style data, and general natural-scene data. The document-heavy subset includes DocVQA~\cite{mathew2021docvqa}, ChartQA~\cite{masry2022chartqa}, InfographicVQA~\cite{mathew2022infographicvqa}, and ScienceQA~\cite{lu2022scienceqa}, which provide recurring layouts such as typography, grids, tables, axes, legends, and OCR structures. The spatial-structure subset includes Visual Genome~\cite{krishna2017visualgenome}, GQA~\cite{hudson2019gqa}, and RefCOCO/RefCOCOg~\cite{yu2016refcoco}, which emphasize object relations, compositional reasoning, and referring expressions. The diagram subset includes AI2D~\cite{kembhavi2016ai2d}, while the general natural-scene subset includes AOKVQA~\cite{schwenk2022aokvqa} and TextCaps~\cite{sidorov2020textcaps}.

% For ScienceQA during mid-training, we use a different target construction policy. Since ScienceQA provides lecture annotations in addition to questions and answers, we use the lecture followed by the answer as the prediction target. This provides reasoning-oriented explanatory supervision and encourages the model to associate visual memory retrieval with intermediate semantic context. During SFT, we do not use this policy and instead train the model to predict the final answer.

In the mid-training stage, we organize the data into four groups to help the model learn various visual patterns. The first group focuses on documents and charts, using DocVQA~\cite{mathew2021docvqa}, ChartQA~\cite{masry2022chartqa}, InfographicVQA~\cite{mathew2022infographicvqa}, and ScienceQA~\cite{lu2022scienceqa} to learn recurring layouts such as tables, grids, and text structures. The second group includes spatial-structure data, Visual Genome~\cite{krishna2017visualgenome}, GQA~\cite{hudson2019gqa}, and RefCOCO/RefCOCOg~\cite{yu2016refcoco}, which help the model understand how objects relate to each other. The remaining groups consist of diagrams from AI2D~\cite{kembhavi2016ai2d} and natural scenes from AOKVQA~\cite{schwenk2022aokvqa} and TextCaps~\cite{sidorov2020textcaps}. For ScienceQA during this stage, we use a special method where the model predicts the lecture explanation before the answer. This helps the model connect its visual memory with the underlying meaning of the image.

% The goal of SFT is reasoning alignment after memory retrieval. Therefore, we use high-reasoning datasets, including ScienceQA, AI2D, DocVQA, ChartQA, GQA, and AOKVQA. This stage aligns the retrieved memory representations with downstream response generation, encouraging the model to use retrieval as support for deeper visual and multimodal reasoning.

During SFT, we use datasets that require high-level thinking, including ScienceQA, AI2D, DocVQA, ChartQA, GQA, and AOKVQA, to teach the model how to use retrieved information correctly. During this stage, we no longer use the lecture explanation policy and instead train the model to predict the final answer directly. This process ensures the model can use its memory to support deeper reasoning and generate accurate responses for complex tasks.

For evaluation-domain adaptation, we additionally use 10$\%$ of each evaluation dataset as few-shot training data and reserve the remaining 90$\%$ strictly for evaluation.

\input{tables/main_table}

% \paragraph{Backbone models and Baselines.}
% We use Qwen3-VL-2B-Instruct \cite{bai2025qwen3} as the backbone model for all experiments and evaluate \textsc{V-Lumen} under three baseline settings designed to isolate different sources of improvement. First, we compare the frozen vanilla backbone against the same backbone augmented with \textsc{V-Lumen}, which measures the effect of adding lookup-based visual memory without changing the underlying VLM. Second, we consider strong LoRA adaptation baselines. We compare \textsc{V-Lumen}(1)+LoRA(k) against LoRA(k), which controls for the backbone adaptation budget, and against LoRA(1+k), which matches the total trainable-parameter budget of \textsc{V-Lumen} plus LoRA using LoRA alone. This separates gains from \textsc{V-Lumen}’s memory organization from gains that could be explained simply by adding more trainable adapter parameters. Third, we compare against an MoE-based sparse-capacity baseline. We construct MoE$(1+k)$ to approximately match \textsc{V-Lumen}$(1)$+MoE$(k)$ in both accessible parameters and activated parameters, making the comparison reflect \textsc{V-Lumen}’s central design goal: expanding representational capacity while keeping inference-time activated computation bounded. Together, these baselines cover frozen-backbone augmentation, dense low-rank adaptation, and sparse conditional computation, allowing us to evaluate whether \textsc{V-Lumen} improves performance through its lookup-based capacity mechanism rather than through parameter count or adaptation alone.

\paragraph{Evaluations.}
We evaluated \textsc{V-Lumen} on five vision--language benchmarks that emphasize reasoning: ChartMuseum \cite{tang2025chartmuseum}, FRIEDA \cite{pyo2025frieda}, VisRes \cite{tortei2025visres}, MMMU \cite{yue2024mmmu}, and MMMU-Pro \cite{yue2025mmmu}. 
% ChartMuseum and VisRes were evaluated on their test splits, FRIEDA on its data split, MMMU on the validation split, and MMMU-Pro on the test split of the standard configuration. 
For consistency with our few-shot protocol, evaluation used the held-out 90\% partition induced by the fixed split seed, from which we sampled up to 1,000 examples for each benchmark.
We report relaxed accuracy: for numeric answers we accept predictions within 5\% relative error of the ground truth when both can be parsed as numbers; otherwise we compare predictions to answers after light normalization (case, punctuation, whitespace). 
% For multiple-choice tasks (e.g., MMMU / MMMU-Pro), we extract the leading option letter from the model’s free-form output before matching.

\paragraph{Training details.}
% \textsc{\textsc{V-Lumen}} is trained in two stages. 
% Both stages updates only the trainable paths of each configuration (LoRA/MoE weights and/or \textsc{V-Lumen} tables + fusion parameters) while freezing the backbone $f_{\theta}$. The training objective is the next-token cross-entropy under an answer-only mask(image and prompt tokens do not contribute to the loss). 
% for \textsc{V-Lumen}{+}LoRA/MoE, LoRA and the fusion parameters are fine-tuned jointly. Both stages use AdamW ($\beta{=}(0.9, 0.95)$, weight decay $0.01$), bf16 mixed precision, a cosine learning-rate schedule with a $200$-step linear warmup, and gradient clipping at global norm $1.0$.
\textsc{V-Lumen} is trained in two stages.
In both stages, we update only the trainable paths of each configuration, including LoRA/MoE weights and/or \textsc{V-Lumen} memory tables and fusion parameters, while keeping the backbone $f_{\theta}$ frozen.
The training objective is next-token cross-entropy with an answer-only mask, where image and prompt tokens do not contribute to the loss.
For \textsc{V-Lumen}{+}LoRA/MoE, the adaptation modules and \textsc{V-Lumen} fusion parameters are fine-tuned jointly.
Both stages use AdamW ($\beta{=}(0.9, 0.95)$, weight decay $0.01$), bf16 mixed precision, a cosine learning-rate schedule with a $200$-step linear warmup, and gradient clipping at global norm $1.0$. For additional implementation details, please refer to Appendix.~\ref{app:impl}.
% 

% \paragraph{Implementation details.}\textcolor{blue}{ToDo}

% \input{tables/main_table}
% \input{tables/ablation_table}

\subsection{Experimental Results}

\paragraph{Reasoning-intensive vision-language benchmarks.}
% Tab.~\ref{tab:main_results} reports results on five reasoning-intensive vision-language benchmarks. The three comparison blocks answer different questions because they impose different \emph{activation} properties on the added capacity. In the LoRA-based block, the baseline is a dense adapter, in which essentially all newly introduced trainable parameters participate in every token's forward pass; matching the budget with LoRA($1{+}k$) therefore increases both trainable and \emph{per-token activated} capacity. The vanilla- and MoE-based settings instead follow a lookup-style regime: most parameters are \emph{accessible}, but only a routed subset is \emph{activated} per token. This matches the objective of \textsc{V-Lumen}, which expands externally stored capacity without treating every added parameter as dense per-token computation.
Tab.~\ref{tab:main_results} reports results on five reasoning-intensive vision-language benchmarks, using Qwen3-VL-2B-Instruct~\cite{bai2025qwen3} as the backbone for all experiments. The table consists of three comparison blocks. 
(i) The \emph{vanilla-based} block attaches \textsc{V-Lumen} to the frozen backbone and asks whether a visual lookup memory alone improves reasoning. 
% (ii) The \emph{LoRA-based} block compares against a dense adapter at a matched trainable budget; in a dense adapter, every added parameter participates in each token's forward pass. (iii) The \emph{MoE-based} block compares against a sparse MoE at matched accessible and activated budgets; like \textsc{V-Lumen}, an MoE keeps many parameters accessible but activates only a subset per token.
(ii) The \emph{LoRA-based} block compares LoRA + \textsc{V-Lumen}, which combines LoRA($k$) with \textsc{V-Lumen}(1), against LoRA-only, a larger LoRA($1{+}k$) adapter scaled to the same \emph{total trainable-parameter budget}. The two differ in how this budget is used at inference: in LoRA-only, every added parameter is activated in each token's forward pass, whereas in LoRA + \textsc{V-Lumen}, only the LoRA component is fully activated and the memory contributes just the few rows retrieved for each visual token.
(iii) The \emph{MoE-based} block compares MoE + \textsc{V-Lumen}, which combines MoE($k$) with \textsc{V-Lumen}(1), against MoE-only, a larger MoE($1{+}k$) matched in both \emph{accessible} parameters (all parameters the model can route to) and \emph{activated} parameters per token. Both sides are sparse: MoE-only activates the routed expert for each token, while MoE + \textsc{V-Lumen} activates a smaller routed expert together with the retrieved memory rows.

\textbf{\emph{Frozen backbone.}} Attaching \textsc{V-Lumen} to the frozen vanilla backbone improves accuracy by $+17.79 \pm 0.47$pp on average, with gains on all five benchmarks. Without updating any backbone weight, adding a visual lookup memory alone substantially strengthens visual reasoning. This setting uses a longer training schedule than the adapter-based settings (Appendix~\ref{app:training}), so we do not compare it directly against LoRA-only or MoE-only.

\textbf{\emph{Matched budgets.}} When part of the budget is reallocated from adapter parameters to \textsc{V-Lumen}, performance is preserved. In the MoE-based block, MoE + \textsc{V-Lumen} reaches $+16.92 \pm 0.91$pp versus $+16.85 \pm 0.95$pp for MoE-only, with higher accuracy on ChartMuseum and VisRes and lower accuracy on the other three benchmarks. In the LoRA-based block, LoRA + \textsc{V-Lumen} reaches +21.15 $\pm$ 0.02pp versus
+20.58 $\pm$ 0.85pp for LoRA-only (+0.57pp), higher on MMMU-Pro and ChartMuseum,
on par on FRIEDA and MMMU, and lower on VisRes. Both differences are within about
one seed standard deviation of the adapter-only baseline, so we do not claim that
lookup memory improves over adaptation at a fixed budget. Rather, these results show that \textsc{V-Lumen} acts as an \emph{independent capacity axis}: capacity can be moved from adapter parameters into a sparsely addressed memory with little or no loss in accuracy at a matched budget, and this memory scales without a corresponding increase in per-token compute (Sec.~\ref{sec:retrieval}).

\paragraph{Zero-shot evaluation.}
We use a 10\% few-shot / 90\% evaluation protocol on the benchmarks in Tab.~\ref{tab:main_results}. These benchmarks are recent and highly challenging, and the 2B backbone scores near zero without in-domain adaptation, so a small few-shot split is needed to obtain a measurable signal.
To evaluate \textsc{V-Lumen} without any in-domain adaptation, we further test it in a fully zero-shot setting on three benchmarks outside our training curriculum, comparing the frozen Qwen3-VL-2B backbone with \textsc{V-Lumen} disabled and enabled on 1,000 samples per benchmark (Tab.~\ref{tab:zeroshot}).
None of the three benchmarks shares image sources with our training data (Appendix~\ref{app:overlap}), and no in-domain few-shot samples are used.
\textsc{V-Lumen} improves over the frozen backbone on all three benchmarks and in every seed, spanning three independently sourced image domains. This suggests that the retrieved memory encodes reusable visual priors rather than memorizing the training distribution.

\input{tables/zeroshot_table}

%% file: tables/main_table.tex
\begin{table}[t]
\centering
\caption{Accuracy (\%) on five reasoning-intensive vision-language benchmarks, reported as mean $\pm$ std over three seeds. Avg $\Delta$ is the average gain over the vanilla backbone (pp). Within each LoRA and MoE pair, both settings are matched in total parameter budget (Appendix~\ref{app:budget}): LoRA-only and MoE-only correspond to LoRA($1{+}k$) and MoE($1{+}k$), and the $+$\textsc{V-Lumen} variants combine LoRA($k$) or MoE($k$) with \textsc{V-Lumen}(1), where $k = 0.5$. Training schedules are listed in Appendix~\ref{app:training}.}
\label{tab:main_results}
\footnotesize
\setlength{\tabcolsep}{3pt}
\begin{tabular}{l ccccc >{\columncolor{gray!10}}c >{\columncolor{gray!10}}c}
\toprule
Setting & ChartMus. & Frieda & VisRes & MMMU & MMMU-Pro & Avg & Avg $\Delta$ \\
\midrule
Vanilla & 1.69 & 2.27 & 5.00 & 1.63 & 0.30 & 2.18 & -- \\
\midrule
\textsc{V-Lumen}-only
  & 16.70{\tiny$\pm$0.79} & 4.77{\tiny$\pm$0.23} & 24.50{\tiny$\pm$2.07} & 33.38{\tiny$\pm$0.95} & 20.50{\tiny$\pm$0.82} & 19.97{\tiny$\pm$0.47} & +17.79{\tiny$\pm$0.47} \\
% \midrule
% LoRA-only
  % & 18.24{\tiny$\pm$0.70} & 5.53{\tiny$\pm$0.69} & 23.63{\tiny$\pm$1.40} & 43.37{\tiny$\pm$0.82} & 28.20{\tiny$\pm$1.21} & 23.79{\tiny$\pm$0.33} & +21.62{\tiny$\pm$0.33} \\
\midrule
LoRA-only
  & 17.87{\tiny$\pm$0.28} & 5.53{\tiny$\pm$0.13} & 22.03{\tiny$\pm$0.50} & 41.87{\tiny$\pm$1.47} & 26.50{\tiny$\pm$1.93} & 22.76{\tiny$\pm$0.85} & +20.58{\tiny$\pm$0.85} \\
LoRA + \textsc{V-Lumen}
  & 18.24{\tiny$\pm$0.41} & 5.61{\tiny$\pm$0.35} & 21.70{\tiny$\pm$2.39} & 41.91{\tiny$\pm$1.30} & 29.20{\tiny$\pm$0.66} & 23.33{\tiny$\pm$0.02} & +21.15{\tiny$\pm$0.02} \\
\midrule
MoE-only
  & 13.48{\tiny$\pm$0.90} & 4.85{\tiny$\pm$0.35} & 23.53{\tiny$\pm$3.87} & 33.21{\tiny$\pm$2.22} & 20.10{\tiny$\pm$0.26} & 19.03{\tiny$\pm$0.95} & +16.85{\tiny$\pm$0.95} \\
MoE + \textsc{V-Lumen}
  & 14.19{\tiny$\pm$0.49} & 4.70{\tiny$\pm$1.02} & 25.07{\tiny$\pm$1.43} & 32.54{\tiny$\pm$1.17} & 19.00{\tiny$\pm$2.59} & 19.10{\tiny$\pm$0.91} & +16.92{\tiny$\pm$0.91} \\
\bottomrule
\end{tabular}
\end{table}

%% file: tables/zeroshot_table.tex
\begin{table}[t]
\centering
\caption{Zero-shot accuracy (\%) on benchmarks outside the training curriculum, using the frozen Qwen3-VL-2B backbone (mean $\pm$ std over three seeds). No in-domain few-shot samples are used, and none of the benchmarks shares image sources with our training data (Appendix~\ref{app:overlap}). Gains over the vanilla backbone are shown in \textcolor{green!45!black}{green}.}
\label{tab:zeroshot}
\begin{tabular}{lcl}
\toprule
Benchmark & Vanilla & +\textsc{V-Lumen} \\
\midrule
TextVQA~\cite{singh2019towards}    & 11.00 & 37.17{\scriptsize$\pm$3.42} {\scriptsize\textcolor{green!45!black}{(+26.17)}} \\
OCR-VQA~\cite{mishra2019ocr}       & 13.80 & 47.67{\scriptsize$\pm$0.76} {\scriptsize\textcolor{green!45!black}{(+33.87)}} \\
VizWiz-VQA~\cite{gurari2018vizwiz} &  2.00 & \phantom{0}9.97{\scriptsize$\pm$1.84} {\scriptsize\textcolor{green!45!black}{(+7.97)}} \\
\bottomrule
\end{tabular}
\end{table}

%% file: sections/5_analysis.tex
\section{Analysis}
\label{sec:analysis}

We use the logit lens to examine where in depth \textsc{V-Lumen} changes the model's prediction. For each decoder layer $\ell$, we project the hidden state of the last token through the final $\mathrm{RMSNorm}$ and the tied LM head, obtaining a distribution $p_\ell$, and measure (i) the Jensen--Shannon divergence $\mathrm{JSD}(p_\ell, p_{\text{final}})$ to the model's own final prediction, computed with the natural logarithm (in nats), and (ii) the top-1 accuracy of $p_\ell$ against the gold answer. No probe parameters are trained. Text-only Engram~\cite{cheng2026conditional} reports that lookup memory lets intermediate layers approach the final prediction earlier, which it interprets as accelerated convergence; we test whether visual memory behaves similarly.

\begin{figure}[t]
\centering
\includegraphics[width=0.7\linewidth]{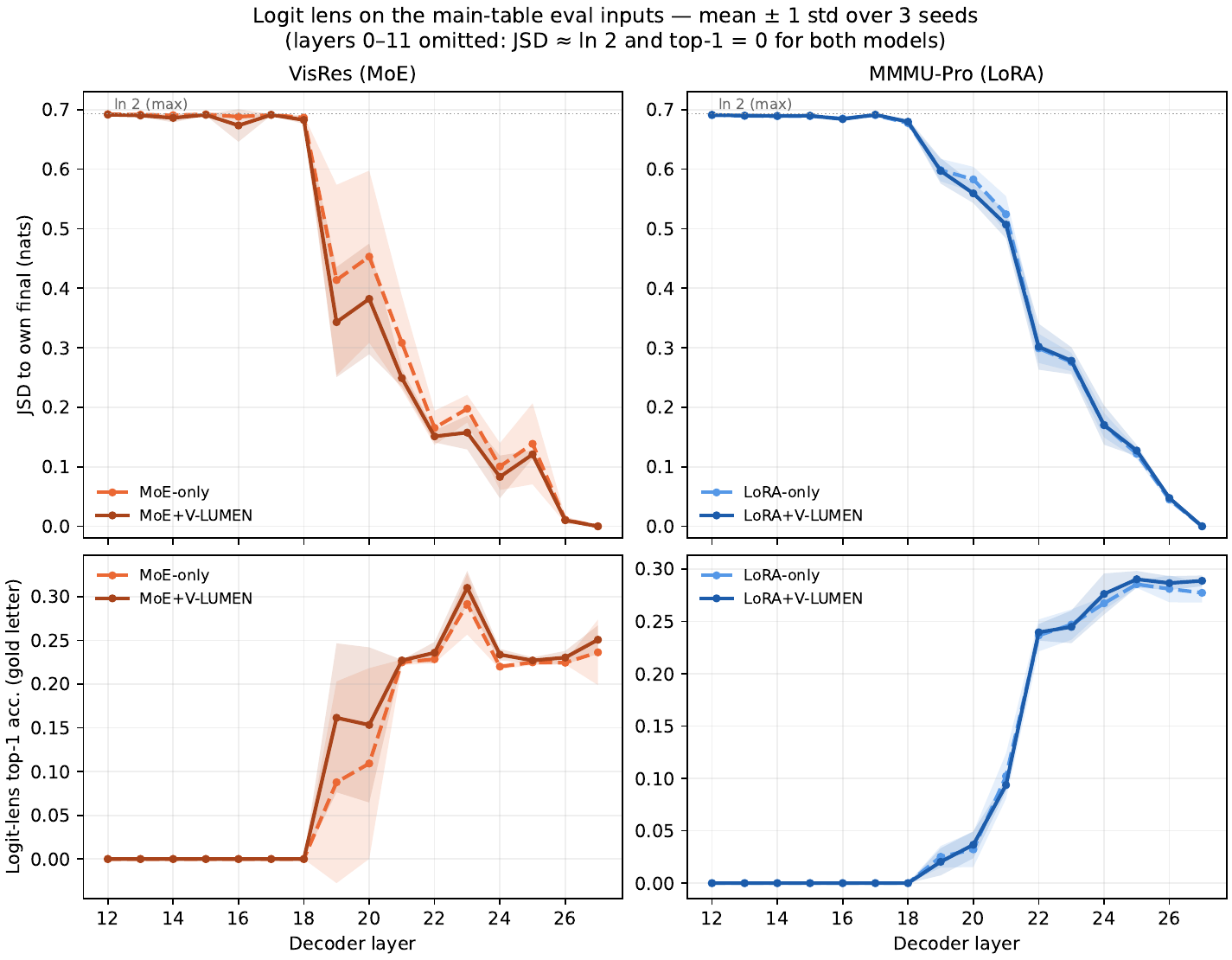}
% \caption{Layer-wise logit-lens analysis on VisRes (MoE, left) and MMMU-Pro (LoRA, right), comparing each adapter-only model with its budget-matched $+$\textsc{V-Lumen} counterpart. Top: $\mathrm{JSD}(p_\ell, p_{\text{final}})$ to the model's own final prediction. Bottom: top-1 accuracy of the gold answer. Layers 12--27 are shown; below layer 19, both models stay at the maximum JSD and zero top-1 accuracy. Mean $\pm$ 1 std over three seeds.}
\caption{Layer-wise logit-lens analysis on VisRes (MoE, left) and MMMU-Pro (LoRA,
right), comparing each adapter-only model with its +\textsc{V-Lumen} counterpart (MoE:
budget-matched pair of Tab.~\ref{tab:main_results}; LoRA: rank-matched LoRA(0.5)
pair of Tab.~\ref{tab:lora_design1}). Top: JSD$(p_\ell, p_{\text{final}})$ to the
model's own final prediction. Bottom: top-1 accuracy of the gold answer. Layers
12--27 are shown; below layer 19, both models stay at the maximum JSD and zero
top-1 accuracy. Mean $\pm$ 1 std over three seeds.}
\label{fig:logit_lens}
\end{figure}

\paragraph{Setup.}
% We probe the main-table checkpoints of Tab.~\ref{tab:main_results} (fusion layers (2, 8)), using the same inputs and evaluation samples as the main evaluation. Among the four (adapter, benchmark) pairs in Tab.~\ref{tab:main_results} where adding \textsc{V-Lumen} improves over the matched adapter-only model, we analyze the two whose answers are a single option letter, VisRes (MoE) and MMMU-Pro (LoRA), for which the layer-wise top-1 accuracy of the gold answer is well defined. Since these pairs are selected because \textsc{V-Lumen} improves on them, they are not a random sample of benchmarks. All results are means over three seeds (Fig.~\ref{fig:logit_lens}).
We probe the main-table checkpoints (fusion layers (2, 8)), using the same inputs and evaluation samples as the main evaluation. For MoE, we use the budget-matched pair of Tab.~\ref{tab:main_results}. For LoRA, we use the rank-matched pair of Tab.~\ref{tab:lora_design1} (LoRA(0.5) with and without \textsc{V-Lumen}), whose adapter is identical on both sides, so that layer-wise differences isolate the effect of the memory. Among the four (adapter, benchmark) pairs in which adding \textsc{V-Lumen} improves over the corresponding adapter-only model, we analyze the two whose answers are a single option letter, VisRes (MoE) and MMMU-Pro (LoRA), for which the layer-wise top-1 accuracy of the gold answer is well defined. Since these pairs are selected because \textsc{V-Lumen} improves on them, they are not a random sample of benchmarks. All results are means over three seeds (Fig.~\ref{fig:logit_lens}).

\paragraph{\textsc{V-Lumen} tends to approach the final prediction earlier.}
Up to layer 18, neither model ranks the gold answer first and the JSD stays at its maximum ($\ln 2$) on both benchmarks, so the probes can distinguish the two models only from layer 19 onward. There, the \textsc{V-Lumen} models tend to approach their final prediction earlier. Averaged over layers 19--26, the JSD to the final prediction is lower with \textsc{V-Lumen} on both benchmarks ($-0.036$ on VisRes and $-0.004$ on MMMU-Pro), and on VisRes the gold answer becomes the top-1 prediction earlier, with leads of $+7.4$ and $+4.4$ points at layers 19 and 20 in two of three seeds. At the final layer, top-1 accuracy rises from 0.236 to 0.251 on VisRes and from 0.277 to 0.289 on MMMU-Pro, matching the margins in Tab.~\ref{tab:main_results}. This pattern is consistent with the accelerated convergence reported for Engram. However, the differences are small relative to seed variance: in each pair, only two of three seeds show a lower JSD, and the mean difference lies within one standard deviation across seeds. We therefore read these probes as a tendency rather than a consistent effect, in line with the modest accuracy margins in Tab.~\ref{tab:main_results}.

% Optional (if space allows):
% On ChartMuseum (MoE), whose open-ended answers we analyze with JSD only, we additionally observe a small but seed-consistent increase in JSD around layer 16 ($+0.012$ in all three seeds), which may reflect a perturbation of the intermediate trajectory by the retrieved memory; FRIEDA (LoRA) shows no such increase (Appendix~\ref{app:logit_lens}).

%% file: sections/6_relworks.tex
\section{Related Work}
\label{sec:related}

% \subsection{Embedding Scaling in LLMs}
% Recent work ~\cite{cheng2026conditional, sadhukhan2026stem, yu2025scaling, jie2025mole, gemma3n, roy2022ngrammer} has explored embedding-level scaling as an alternative capacity axis for language models.
% Lookup-based methods such as Engram~\cite{cheng2026conditional} and STEM~\cite{sadhukhan2026stem} expand accessible capacity through sparsely accessed embedding memories, while related approaches scale token or $n$-gram representations through latent or precomputed embeddings~\cite{roy2022ngrammer,yu2025scaling}.
% These methods benefit from the discrete nature of language tokens, which naturally supports vocabulary expansion, $n$-gram indexing, and deterministic lookup.
% In contrast, visual representations in VLMs are continuous and spatially structured, making direct application of text-based embedding scaling non-trivial.
% \textsc{V-Lumen} extends the lookup-based scaling idea to the visual domain by discretizing continuous visual features and constructing spatially and text-conditioned memory keys for external visual memory retrieval.
% A more detailed discussion is provided in \cref{app:embedding_scaling}.

\subsection{Embedding Scaling in LLMs}
\label{sec:related:embedding}
Embedding-level scaling has emerged as an axis of model capacity that is complementary to scaling depth, width, or dense computation. Lookup-based methods expand accessible capacity through sparsely accessed embedding memories: Engram~\cite{cheng2026conditional} retrieves static $n$-gram memories through deterministic lookup and fuses them into the hidden state, and STEM~\cite{sadhukhan2026stem} replaces FFN up-projections with layer-local embedding lookup, decoupling capacity growth from per-token dense computation. A related line scales token or $n$-gram representations themselves: N-Grammer~\cite{roy2022ngrammer} augments transformers with latent $n$-gram embeddings built from discrete token representations, SCONE~\cite{yu2025scaling} learns contextualized embeddings for frequent $n$-grams with a separate transformer, and vocabulary scaling studies show that larger or compute-optimal vocabularies improve language modeling efficiency and performance~\cite{huang2025overtokenized,tao2024vocabscaling}.

Several of these designs also exploit the fact that embedding tables need not reside in accelerator memory. SCONE precomputes its embeddings into an off-accelerator store, MoLE~\cite{jie2025mole} reparameterizes expert computation as lookup tables for inference, and the per-layer embeddings of Gemma 3n~\cite{gemma3n} can be generated outside the model's operating memory and cached for layer-wise execution. Together, these works treat embedding tables as a capacity reservoir that grows in storage rather than in per-token compute.

All of these methods, however, rely on the discreteness of language tokens, which naturally supports vocabulary expansion, $n$-gram indexing, and deterministic lookup. Visual features in VLMs are instead continuous, spatially structured, and query-dependent, so text-based embedding scaling does not transfer directly. \textsc{V-Lumen} extends lookup-based scaling to this setting by discretizing continuous visual features and constructing spatially aware and text-conditioned memory keys. Related work on spatial aggregation is discussed in Appendix.~\ref{app:related}.

\subsection{Hash-based and Deterministic Routing}
Sparse-MoE architectures route inputs via a learned gate, but the learned router is notoriously unstable: load-imbalance collapse and auxiliary balancing losses are documented in Switch Transformer~\cite{fedus2022switch}. A parallel line replaces the learned router with a non-parametric function: Hash Layers~\cite{roller2021hash} (fixed hash), BASE Layers~\cite{lewis2021base} (balanced assignment as a linear program), and THOR~\cite{zuo2022thor} (stochastic expert selection). Our TCH (\S\ref{sec:method:retrieval}) transposes this pattern from expert selection to memory addressing: the same patch under different queries is routed to different slots of one memory bank via a frozen random projection and a rank-bucket XOR.

%% file: sections/7_conclusion.tex
% 추가할 내용
% Although \textsc{V-Lumen} is structurally compatible with offloading memory storage outside accelerator memory, we leave a full systems-level analysis of memory placement, bandwidth, and latency trade-offs to future work.

\section{Conclusion}
% \textcolor{magenta}{
% We introduced \textsc{V-Lumen}, which extends a frozen vision--language backbone with an external visual memory along an axis that is complementary to parameter adaptation: memory addresses are non-parametric, memory contents are learned, and the retrieved vector is fused back into the backbone through a gated residual. Three ingredients make the design work on the vision side of a VLM: discretization through a frozen codebook that fixes the visual token (\S\ref{sec:method:memory}), weight-free spatial aggregation that turns isolated patch codes into region-aware keys (\S\ref{sec:method:spatial}), and text-conditioned hashing that routes the same patch to disjoint memory slots depending on the query (\S\ref{sec:method:retrieval}). 
% }
We introduced \textsc{V-Lumen}, an external visual lookup memory for vision-language models. Three components make lookup-based scaling work on the vision side: a frozen codebook that maps continuous patch features to discrete codes that serve as memory addresses (\S\ref{sec:method:memory}), non-parametric spatial aggregation that turns isolated patch codes into region-aware keys (\S\ref{sec:method:spatial}), and text-conditioned hashing that routes the same patch to different memory slots depending on the query (\S\ref{sec:method:retrieval}). \textsc{V-Lumen} improves a frozen backbone and transfers zero-shot to unseen image domains. At the same total budget, it matches LoRA and MoE adapters while holding part of their capacity in an external memory rather than in adapter parameters, forming an independent capacity axis whose capacity grows in storage rather than in per-token compute. Because this memory is addressed by hashing and read sparsely, it opens a path toward serving visual capacity outside GPU memory.

% \textbf{Future work} 
\section{Future Work}
\label{sec:future}
\paragraph{Serving the memory outside GPU memory.}
Each visual token reads only $|O| \cdot H$ rows of the memory table (Eq.~\eqref{eq:visual-engram-read}), and these rows are addressed by hashing rather than used in dense matrix multiplications. The table can therefore, in principle, reside in host (CPU) memory or SSD-backed storage and be fetched on demand, as in the large embedding tables of industrial recommendation models~\cite{naumov2019dlrm,agarwal2023bagpipe,wilkening2021recssd} and text-side lookup memories~\cite{cheng2026conditional}. Unlike text-side memories, whose addresses depend only on input token IDs, the addresses of \textsc{V-Lumen} are computed from hidden states at the first fusion layer, so fetching for later fusion layers can overlap with computation but fetching for the first cannot. Measuring the latency and throughput of such a deployment, and designing prefetching policies around this constraint, is a natural next step.

\paragraph{Faster key construction.}
Key construction accounts for about three quarters of the prefill overhead in our implementation (Appendix~\ref{app:throughput:prefill}). It consists of integer hashing and indexing over the visual-token grid, so vectorizing it on the GPU while keeping the computed addresses identical should remove most of this overhead.

% \paragraph{\textcolor{magenta}{[TODO: confirm the final results] Turning memory rows into accuracy}} 
% Additional rows are free in per-token compute but do not yet improve accuracy in our setting. Understanding what the memory learns, and how training data and key design affect how much of a larger table is used, is a natural next step.
% TODO (after A7): describe how accuracy changes with the number of rows and what remains open.
\paragraph{Turning memory rows into accuracy.}
Additional rows are free in per-token compute but do not yet improve accuracy in our setting, whereas wider memory vectors do (Appendix~\ref{app:capacity}). Understanding what the memory learns, and how training data and key design affect how much of a larger table is used, could let the row axis deliver the gains that width currently provides without increasing the per-token read.

\paragraph{Broader settings.}
Extending \textsc{V-Lumen} to backbones beyond 2B parameters would test whether the capacity axis persists with scale. The component ablation (Appendix~\ref{app:ablation}) also suggests evaluating settings where visual codes shift systematically, such as reusing a trained memory after the vision encoder or connector is updated.
% [TODO] 실험 후 수정 필요
% Across the evaluation suite of \S\ref{sec:experiments}, \textsc{[V-Lumen]} improves over the frozen backbone and matches or exceeds LoRA-style parameter adaptation under matched compute, while keeping the backbone untouched.

%% file: sections/8_appendix.tex
\section{Limitations}
\label{app:limitations}
% Two limitations scope the claims in the main text. First, the improvements reported in \S\ref{sec:experiments} are \emph{accuracy} gains at a fixed inference configuration. Although \textsc{V-Lumen}'s per-position read (\cref{eq:visual-engram-read}) touches only $|\mathcal{O}|{\cdot}H$ embedding rows---sparse gathers rather than the dense matmul that dominates LoRA and MoE---we do not report wall-clock latency, end-to-end throughput, or measurements of the external memory served from offloaded storage (e.g., CPU-pinned DRAM or NVMe-backed tables). The structural suitability of the external memory for hierarchical memory serving is therefore an argument from design, not an empirical result, and a direct throughput/latency characterisation is left to future work. Second, all experiments use a single backbone scale, Qwen3-VL-2B-Instruct. Whether the capacity added by \textsc{V-Lumen} continues to be complementary to backbone adaptation (LoRA/MoE) at larger backbone scales, and how the optimal memory table size and fusion depth shift with backbone capacity, is not verified here. The interpretability findings in \S\ref{sec:analysis} inherit this scope: the observed mid-stack reorganization is established for a 2B backbone, and its scaling behaviour is an open question.
At a matched parameter budget, \textsc{V-Lumen} matches LoRA and MoE adapters in accuracy, so the advantage of holding capacity in an external memory lies in system-level properties that we have argued from its design but not yet fully demonstrated. First, because the memory is addressed by hashing and only a few rows are read per visual token, it can in principle be served from host memory or storage rather than GPU memory. We have not measured such a deployment: in our implementation, the memory table resides in GPU memory in fp32, and the prefill overhead (19--25\,ms; Appendix~\ref{app:throughput}) is dominated by key construction computed with a Python loop on the CPU, so the reported latency reflects our implementation rather than a lower bound. Second, the number of memory rows can grow without increasing per-token compute, as lookup latency stays constant across a $40\times$ range of table sizes (Tab.~\ref{tab:retrieval_latency}). In our current setting, however, additional rows have not yet translated into higher accuracy, whereas widening the retrieved vectors does improve accuracy while slightly increasing the per-token read (Appendix~\ref{app:capacity}). This suggests that the training data and key design, rather than table size, currently bound how much of the memory is used, which leaves room for improving how a larger table is filled and read. Turning these properties into measured benefits is the main direction for future work (\S\ref{sec:future}).

%-----------------------------------------------------------------------------
% \subsection{Future work}
% Because each per-position read in \textsc{V-Lumen} touches only $|\mathcal{O}|{\cdot}H$ embedding rows (\cref{eq:visual-engram-read})---sparse gathers rather than the dense matmul that ties LoRA/MoE adapters to GPU HBM---the external visual memory $\{\mathbf{E}^{\text{vis}}_{n,h}\}$ can in principle be served from CPU-pinned DRAM or NVMe-backed storage, the same memory-hierarchy regime that powers industrial-scale recommendation systems~\cite{naumov2019dlrm,agarwal2023bagpipe,wilkening2021recssd}. Turning this structural property into a measured capability---via throughput/latency characterisation and prefetching policies tuned to hash-address locality---is a natural next step, and opens \textsc{V-Lumen} to larger backbones, more diverse visual domains, and longer instruction-tuned contexts along the same capacity-scaling axis.

\section{Broader Impact}
\label{app:broader_impact}
\textsc{V-Lumen} adds capacity to a vision-language model through an external memory that is addressed by hashing and read only a few rows at a time, rather than through parameters used in every forward pass. If this memory is served from slower but larger memory tiers, such as host memory on a server or flash storage on a device, capacity could be added to VLMs deployed under tight accelerator-memory budgets, including on-device and edge settings, without a matching increase in accelerator memory or per-token compute. Related designs in language models already keep embedding tables outside the accelerator's working memory~\cite{gemma3n, cheng2026conditional}. Such deployments could lower the hardware cost and energy of running capable multimodal models and make them available on consumer hardware and in settings with limited connectivity, where processing images locally can also keep user data on the device. We have not yet measured such deployments (Appendix~\ref{app:limitations}), so these benefits remain potential.

At the same time, making capable VLMs easier to deploy broadly also lowers the barrier to misuse, for example in surveillance or automated decision-making without human oversight, and carries over the risks of the underlying backbone, including incorrect or overconfident predictions and biases inherited from pretrained models and training data. Because the memory table stores content learned from training images, models trained on sensitive data could retain information from that data. Our work does not introduce new datasets of sensitive content or a deployed system, and any practical use should follow the safety, fairness, and privacy requirements of the target application and include human oversight where errors could have significant consequences.

\section{Extended Related Work}
\label{app:related}

\paragraph{Spatial aggregation.}
Our spatial aggregation design (\S\ref{sec:method:spatial}) draws on prior work that uses axis-aligned operations for efficient spatial context modeling, including strip pooling~\cite{hou2020strippool}, criss-cross attention~\cite{huang2019ccnet}, and axial attention~\cite{wang2020axialdeeplab}. It is also related to explicit positional conditioning methods such as CoordConv~\cite{liu2018coordconv} and CoAtNet~\cite{dai2021coatnet}. Unlike these works, \textsc{V-Lumen} applies these ideas to memory addressing rather than feature transformation: row/column summaries and coordinate-bucket salts construct spatially aware hash keys without trainable parameters.

\section{Method Details}
\subsection{Projected Relevance Hashing: Implementation Details}
\label{app:prh}

This appendix expands on the text-conditioning pipeline of \S\ref{sec:method:retrieval}, where the main text keeps only the final XOR update \cref{eq:tch-xor} and defers score / bucketing details here.

\paragraph{Score.}
Let $\mathbf{h}_{\text{txt}} \in \mathbb{R}^{D}$ be the input hidden state of layer $\ell_{1}$ at the \emph{last text position} and $\mathbf{h}_{\text{vis},i} \in \mathbb{R}^{D}$ ($i \in \{1,\dots,N\}$, $N = H_{v} W_{v}$) the image-patch hidden states at the same layer. With two frozen random projections $\mathbf{W}^{Q}_{\text{prh}}, \mathbf{W}^{K}_{\text{prh}} \in \mathbb{R}^{P\times D}$ ($P{=}64$, entries initialized $\mathcal{N}(0, 0.02^{2})$ and never updated), we compute the per-patch score
\begin{equation}
  r^{\text{prh}}_{i} \;=\; \frac{\bigl\langle \mathbf{W}^{Q}_{\text{prh}} \mathbf{h}_{\text{txt}},\, \mathbf{W}^{K}_{\text{prh}} \mathbf{h}_{\text{vis},i}\bigr\rangle}{\sqrt{P}}, \qquad i \in \{1,\dots,N\}.
  \label{eq:app-prh-score}
\end{equation}

\paragraph{Rank-based bucketing.}
Raw scores $r^{\text{prh}}_{i}$ are sensitive to magnitude / sign drift across samples. We therefore rank-bucket them within each image so that only the \emph{ordering} of patches enters the hash key:
\begin{equation}
  b_{i} \;=\; \Bigl\lfloor \frac{\mathrm{rank}(r^{\text{prh}}_{i})\cdot K}{N} \Bigr\rfloor, \qquad b_{i} \in \{0, \dots, K-1\},
  \label{eq:app-rank-bin}
\end{equation}
with $K{=}8$. The resulting grid $\mathbf{B} \in \{0,\dots,K{-}1\}^{H_{v}\times W_{v}}$ is the object folded into the hash key by \cref{eq:tch-xor}.

\paragraph{Why frozen random projections are sufficient.}
(i)~Both \cref{eq:app-rank-bin} and the integer XOR of \cref{eq:tch-xor} are non-differentiable, so end-to-end gradient cannot reach $\mathbf{W}^{Q,K}_{\text{prh}}$ and making the projections ``learnable'' in the standard sense would require a straight-through estimator or a soft-lookup relaxation that reintroduces the learned-router instability the design of \S\ref{sec:method:retrieval} set out to avoid. (ii)~Random Gaussian projections approximately preserve pairwise inner products~\cite{johnson1984jl}: for any $\epsilon>0$ and $m \ge \Theta(\log N / \epsilon^{2})$, a random $\mathbf{W}\!\in\!\mathbb{R}^{m\times D}$ satisfies $|\langle \mathbf{W}\mathbf{u}, \mathbf{W}\mathbf{v}\rangle - \langle \mathbf{u}, \mathbf{v}\rangle| \le \epsilon \|\mathbf{u}\|\|\mathbf{v}\|$ for all $\mathbf{u}, \mathbf{v}$ in a fixed $N$-element set with high probability. With $N{=}64$ patches, even the small $P{=}64$ we use suffices to separate semantically distinct queries into different rank orderings without any training.

%-----------------------------------------------------------------------------
\subsection{Debiased Relevance Hashing (DRH) Details}
\label{app:drh}

DRH (\S\ref{sec:method:retrieval}) replaces only the score in \cref{eq:app-prh-score}; the bucketing and XOR stages are identical. We use the backbone's raw layer-$\ell_{1}$ text-to-visual similarity $r^{\text{raw}}_{i} = \langle \mathbf{h}_{\text{txt}}, \mathbf{h}_{\text{vis},i}\rangle / \sqrt{D}$ (no projection), normalize into a patch probability $\mathbf{p} = \mathrm{softmax}(\mathbf{r}^{\text{raw}})\in\Delta^{N}$, and debias it with an EMA running prior $\bar{\mathbf{p}} \in \mathbb{R}_{\ge 0}^{N}$ that is maintained across training forwards to suppress patches whose mean similarity mass is always high regardless of the query:
\begin{equation}
  r^{\text{drh}}_{i} \;=\; p_{i} \,-\, \lambda\, \bar{p}_{i}, \qquad \bar{\mathbf{p}}^{(t)} \;=\; \gamma\, \bar{\mathbf{p}}^{(t-1)} \,+\, (1-\gamma)\, \mathrm{sg}[\mathbf{p}^{(t)}],
  \label{eq:app-drh}
\end{equation}
with $\lambda{=}1$ and $\gamma{=}0.99$; $\mathrm{sg}[\cdot]$ denotes the stop-gradient applied before the EMA update so that the prior never back-propagates into the backbone. The prior is maintained as a non-trainable buffer of size $N$, initialized to the uniform prior $\bar{p}^{(0)}_{i} = 1/N$, and is updated only during training-mode forwards. At inference the prior is frozen.

\paragraph{PRH $\oplus$ DRH ensemble.}
The two scores can be combined before \cref{eq:app-rank-bin}. We normalize each by its per-image standard deviation and take a convex combination,
\begin{equation}
  r^{\text{ens}}_{i} \;=\; \eta\cdot \tfrac{r^{\text{prh}}_{i}}{\hat{\sigma}_{\text{prh}}} \,+\, (1-\eta)\cdot \tfrac{r^{\text{drh}}_{i}}{\hat{\sigma}_{\text{drh}}}, \qquad \eta \in [0, 1],
  \label{eq:app-ensemble}
\end{equation}
with $\eta{=}0.5$ as the default; $\eta{=}1$ recovers PRH and $\eta{=}0$ recovers DRH. Per-image normalization is necessary because the two raw scales differ by roughly an order of magnitude ($r^{\text{prh}}$ lives in $\mathbb{R}$ while $r^{\text{drh}}$ is a softmax residual near zero). We use the symbol $\eta$ for the ensemble mixing weight to avoid collision with the MHCG gate $\boldsymbol{\alpha}_{\ell,t}$ of \cref{eq:app-mhcg-gate}.

%-----------------------------------------------------------------------------
\subsection{Spatial Key Refiners}
\label{app:refiners}

\paragraph{Axial pooling fallback.}
\cref{eq:axial-summary} requires the row and column summaries $\rho_{\text{row}}(i)$ and $\rho_{\text{col}}(j)$ to be scalar integers that (i) are identical across visually similar rows and columns (for the intended row/column invariance of \S\ref{sec:method:spatial}) and (ii) are deterministic given the ID content. A majority vote satisfies both: whenever any codebook ID repeats in a row, the modal ID is a stable, content-based summary. With an $8{\times}8$ grid over a $V{=}16{,}384$ codebook, however, a non-trivial fraction of rows and columns is \emph{all-distinct}---no ID repeats, so the mode is undefined and $\operatorname*{arg\,max}$ over the unique-count vector is arbitrary.

Returning $0$, the minimum ID, or a random value in this case would create a \emph{large summary discontinuity}: rows with one repeat and rows with none would be pushed into incomparable integer ranges, corrupting the blend in \cref{eq:axial}. We therefore fall back to a deterministic rolling hash of the row content---specifically, a polynomial rolling hash over the \emph{sorted} ID sequence, which realizes the operator $\mathcal{H}(\cdot)$ used in \cref{eq:axial-summary}:
\begin{equation}
  \mathcal{H}(\mathbf{s}) \;=\; \Bigl(\sum_{x \in \mathrm{sort}(\mathbf{s})} x\cdot \pi_{\text{rh}}^{\mathrm{pos}(x)}\Bigr) \bmod 2^{31}, \qquad \pi_{\text{rh}} = 2^{17}-1.
  \label{eq:app-axial-fallback}
\end{equation}
Three properties justify this choice. \emph{Content invariance}: sorting before hashing makes $\mathcal{H}$ depend only on the \emph{multiset} of IDs in the slice, so two rows (resp.\ columns) with the same ID content hash to the same value regardless of within-slice arrangement---the same invariance the modal branch offers. \emph{Scale compatibility}: both the modal branch and $\mathcal{H}$ return non-negative integers in the same $[0, 2^{31})$ range, so no discontinuity is introduced at the branch boundary when \cref{eq:axial} blends the summary with $\mathbf{G}_{i,j}$. \emph{Collision resistance}: the Mersenne prime $\pi_{\text{rh}} = 2^{17}-1$ gives the polynomial a large multiplicative order modulo $2^{31}$, so semantically distinct all-distinct rows are unlikely to collide.

The column summary $\rho_{\text{col}}(j)$ follows the same definition with rows replaced by columns and uses the same $\pi_{\text{rh}}$. This fallback is a correctness rather than an optimization choice: it is what lets \cref{eq:axial-summary,eq:axial} remain well-defined, order-independent operations on every patch, not just on patches whose row or column happens to contain a repeat.

\paragraph{Salt primes for the XOR overlays.}
\label{app:hash_constants}
Both the coordinate-bias update of \cref{eq:coord} and the PRH / DRH update of \cref{eq:tch-xor} mix a scalar salt times a small integer bin id into the spatial key via XOR. The two salts $\pi_{\text{crd}}, \pi_{\text{tch}}$ are chosen to be (i) \emph{coprime} to each other and (ii) large compared with the number of coordinate bins ($H_p W_p = 16$) and the number of rank buckets ($K{=}8$), so that the products $p(i,j)\cdot \pi_{\text{crd}}$ and $\mathbf{B}_{i,j}\cdot \pi_{\text{tch}}$ occupy different, well-separated regions of the 64-bit integer key and their XOR contributions do not systematically cancel. We use $\pi_{\text{crd}} = 1{,}000{,}003$ and $\pi_{\text{tch}} = 2{,}000{,}003$. The specific pair of primes is not load-bearing — any two coprime primes in the same magnitude class yield the same qualitative behaviour.

%-----------------------------------------------------------------------------
\subsection{Rolling-Hash Index Map for the V-LUMEN Memory Tables}
\label{app:hash}

The memory read \cref{eq:visual-engram-read} writes $\phi_{n,h}(\kappa^{\text{vis}}_{i,j})$ for brevity; we now give its explicit form, which is a polynomial rolling hash over the preceding $n$ cells in row-major order.
Let $t \in \{1, \dots, T\}$ index visual-token positions with $\kappa^{\text{vis}}_{t} \equiv \kappa^{\text{vis}}_{i(t), j(t)} = \widetilde{\mathbf{G}}_{i(t), j(t)}$. For order $n\in\mathcal{O}$ and head $h\in\{1,\dots,H\}$, let $\boldsymbol{\alpha}_{n,h}\in \mathbb{Z}_{>0}^{n}$ be random integer coefficients sampled once from $\mathrm{Uniform}(1, 10^{6})$ and kept fixed throughout training, and let $P_{n,h}$ be a prime per-head table size, chosen distinctly across $(n,h)$ pairs so that independent heads collide on different partitions of memory. The index map at position $t$ is
\begin{equation}
  \phi_{n,h}(t) \;=\; \Bigl(\sum_{k=0}^{n-1} \alpha_{n,h,k}\cdot \kappa^{\text{vis}}_{\,t-n+1+k}\Bigr)\ \bmod\ P_{n,h},
  \label{eq:app-rolling-hash}
\end{equation}
where out-of-sequence positions $(t-n+1+k) < 1$ are zero-padded. The first $n-1$ positions of each sequence are masked out because they lack a complete $n$-gram history. Distinct primes per $(n,h)$ minimize co-collisions across heads, so different heads address different regions of memory for the same $n$-gram. The rolling hash is a drop-in replacement for the deterministic multiplicative-XOR hash of~\cite{cheng2026conditional}; results in our experiments are insensitive to this choice because the only property the memory tables rely on is that the map be fixed and collision-balanced across the index space.

\paragraph{Note on $\alpha$ overloading.}
Within this appendix $\boldsymbol{\alpha}_{n,h}$ denotes the rolling-hash coefficients of \cref{eq:app-rolling-hash}, which are scalar integers and never appear together with the MHCG gate $\boldsymbol{\alpha}_{\ell,t}$ of \cref{eq:app-mhcg-gate}. They live in disjoint scopes (hash vs.\ fusion) and should not be confused.

\subsection{Distinction from Cross-Modal Query Bridges}
\label{app:query_bridge_distinction}

Prior VLMs often use trainable query or bridge modules to connect visual encoders with language models, such as Perceiver-style latent arrays~\cite{jaegle2022perceiverio}, Flamingo's Perceiver Resampler and gated cross-attention~\cite{alayrac2022flamingo}, and the Q-Former modules in BLIP-2~\cite{li2023blip2} and InstructBLIP~\cite{dai2023instructblip}.
Although these methods also introduce text- or query-dependent visual processing, their role is fundamentally different from our text-conditioned hashing.
Bridge modules learn to transform, compress, or select visual features so that they can be consumed by the language model.
In contrast, \textsc{V-Lumen} keeps the backbone interface unchanged and uses text-conditioned signals only to alter the memory addressing process for external visual memory lookup.

Therefore, our method should not be interpreted as another cross-modal connector.
The projected relevance hashing component uses a frozen random projection and similarity-based scoring, rather than a learned query module or soft cross-attention.
Its goal is not to produce language-aligned visual tokens, but to make the memory address depend on the input instruction.
This provides a lightweight distinction from query-bridge architectures: they learn how to read visual features into the language model, whereas \textsc{V-Lumen} determines which reusable memory embeddings are retrieved for a given visual pattern and text query.

\subsection{Conditional Memory Fusion Details}
\label{app:fusion}

This section provides the details of the fusion module summarized in \S\ref{sec:method:fusion}.
Given the retrieved visual memory vector $\mathbf{e}^{\text{vis}}_{t}$, the goal of the fusion module is to convert it into a residual update $\Delta \mathbf{h}_{\ell,t}$ that can be injected into the backbone hidden state at layer $\ell$.
Since $\mathbf{e}^{\text{vis}}_{t}$ is constructed by concatenating multiple memory cells retrieved from different orders and hash heads, we allow each cell to contribute independently rather than applying a single scalar gate to the entire vector.

\paragraph{Multi-head conditional gating.}
Let $C = |\mathcal{O}|H$ denote the number of retrieved memory cells.
We split the retrieved memory vector into $C$ chunks,
\begin{equation}
    \mathbf{e}^{\text{vis}}_{t}
    =
    \operatorname{concat}
    \left(
    \mathbf{e}^{(1)}_{t}, \dots, \mathbf{e}^{(C)}_{t}
    \right),
    \qquad
    \mathbf{e}^{(c)}_{t} \in \mathbb{R}^{d_{\text{head}}}.
\end{equation}
For the hidden state $\mathbf{h}_{\ell,t}\in\mathbb{R}^{D}$ at layer $\ell$, we compute a cell-wise gate conditioned on the current backbone representation:
\begin{equation}
  \boldsymbol{\alpha}_{\ell,t}
  =
  \sigma\!\left(\mathbf{W}_{g,\ell}\mathbf{h}_{\ell,t}\right)
  \in (0,1)^C,
  \label{eq:app-mhcg-gate}
\end{equation}
where $\mathbf{W}_{g,\ell}\in\mathbb{R}^{C\times D}$ is a layer-specific gating matrix.
Each retrieved memory cell is then projected into the backbone hidden dimension and weighted by its corresponding gate:
\begin{equation}
  \mathbf{o}_{\ell,t}
  =
  \sum_{c=1}^{C}
  \alpha_{\ell,t,c}\,
  \mathbf{V}_{\ell,c}\mathbf{e}^{(c)}_{t},
  \label{eq:app-mhcg-output}
\end{equation}
where $\mathbf{V}_{\ell,c}\in\mathbb{R}^{D\times d_{\text{head}}}$ is a layer-specific projection for the $c$-th memory cell.
This design allows different retrieved cells to be selectively amplified or suppressed depending on the current hidden state.

\paragraph{Depthwise-convolution smoothing.}
Before adding the memory update to the residual stream, we apply a zero-initialized depthwise 1-D convolution along the backbone token sequence.
Let
\begin{equation}
    \mathbf{o}_{\ell}
    =
    \left[
    \mathbf{o}_{\ell,1},
    \dots,
    \mathbf{o}_{\ell,T}
    \right]
\end{equation}
denote the fused memory outputs over the full sequence of length $T$, including both text and visual tokens.
We use a depthwise convolution with kernel size $3$:
\begin{equation}
    \widetilde{\mathbf{o}}_{\ell,t}
    =
    \mathbf{o}_{\ell,t}
    +
    \mathrm{DWConv}_{k=3,\ell}(\mathbf{o}_{\ell})_{t}.
    \label{eq:app-dwconv}
\end{equation}
The convolution is initialized to zero so that the fusion module initially behaves as an identity residual path with respect to the convolutional branch.
The spatial structure of \textsc{V-Lumen} is encoded primarily in the hash-key construction described in \S\ref{sec:method:spatial} and \S\ref{sec:method:retrieval}; this convolution only provides lightweight local smoothing along the backbone's native token order.

\paragraph{Residual update.}
The final memory-induced residual update is obtained by normalizing the smoothed fused output:
\begin{equation}
  \Delta \mathbf{h}_{\ell,t}
  =
  \mathrm{RMSNorm}_{\ell}
  \left(
  \widetilde{\mathbf{o}}_{\ell,t}
  \right).
  \label{eq:app-fusion}
\end{equation}
The backbone hidden state is then updated as
\begin{equation}
  \mathbf{h}'_{\ell,t}
  =
  \mathbf{h}_{\ell,t}
  +
  \Delta \mathbf{h}_{\ell,t}.
\end{equation}
In our main experiments, this fusion module is applied only at the selected injection layers $\ell \in \mathcal{L}_{\mathrm{inj}}$.

\section{Implementation Details}
\label{app:impl}
For the main results in Tab.~\ref{tab:main_results}, we set the adaptation budget ratio to $k = 0.5$, where $k$ denotes the trainable adaptation budget relative to the \textsc{V-Lumen} module. In all settings of Tab.~\ref{tab:main_results}, including the LoRA- and MoE-based ones, we inject \textsc{V-Lumen} at layers (2, 8), following the fusion-layer ablation (Appendix~\ref{app:layer_ablation}). Across all experiments, the learnable residual scaling factor for retrieved-memory fusion is initialized to 0.75. When ensembling PRH and DRH, we apply a scale factor of 0.5 to the PRH branch before combining the two hashing signals.

\subsection{Compute Resources}
\label{app:compute}
All training runs use a single NVIDIA RTX 6000 Ada GPU with 48 GB of memory. With the schedules in Tab.~\ref{tab:training_schedule}, one \textsc{V-Lumen}-only run (10k mid-training and 2k SFT steps) takes about 5.5 hours, and one LoRA- or MoE-based run (2k and 1k steps) takes about 1.5 hours. Evaluation on the five benchmarks takes about 2 hours per run. Latency and memory measurements are performed on a single NVIDIA RTX 6000 Ada GPU (Appendix~\ref{app:throughput}). In total, the experiments reported in this paper, including three seeds per setting, required about 700 GPU-hours.

\subsection{Training Schedules}
\label{app:training}

Tab.~\ref{tab:training_schedule} lists the number of training steps for each setting in Tab.~\ref{tab:main_results}. All other training settings are identical across settings, with a batch size of 4 for mid-training and 8 for SFT. Within each LoRA and MoE pair, both settings share the same schedule, so pairwise comparisons are unaffected. The \textsc{V-Lumen}-only setting, built on the frozen vanilla backbone, uses a longer schedule; we therefore do not compare it directly with the LoRA- and MoE-based settings.

\begin{table}[h]
\centering
\caption{Training steps for each setting.}
\label{tab:training_schedule}
\small
\begin{tabular}{lcc}
\toprule
Setting & Mid-training steps & SFT steps \\
\midrule
\textsc{V-Lumen}-only (frozen vanilla backbone) & 10,000 & 2,000 \\
LoRA-only / LoRA + \textsc{V-Lumen}             & 2,000  & 1,000 \\
MoE-only / MoE + \textsc{V-Lumen}               & 2,000  & 1,000 \\
\bottomrule
\end{tabular}
\end{table}

\section{Evaluation Protocol}
\paragraph{Image-level overlap with the training corpus.}
\label{app:overlap}

Zero-shot claims are most often questioned on image exposure rather than annotation leakage. For each of the three zero-shot benchmarks in Tab.~\ref{tab:zeroshot}, the QA pairs are distinct from our training annotations, and we additionally verify that the benchmark shares no images with our training corpus.

\paragraph{TextVQA.}
We evaluate on the TextVQA \cite{singh2019towards} validation split, whose images are drawn from Open Images \cite{kuznetsova2020open}. One of our training sources, TextCaps \cite{sidorov2020textcaps}, reuses TextVQA images and follows the same train/val partition, and we train only on the TextCaps train split. The two sets are therefore disjoint by construction. We further confirm this with an image-ID check, finding that no TextVQA validation image appears in our training data.

\paragraph{VizWiz-VQA and OCR-VQA.}
VizWiz-VQA \cite{gurari2018vizwiz} consists of photos taken by blind and low-vision users through the VizWiz mobile app. OCR-VQA \cite{mishra2019ocr} consists of book-cover images collected from an online bookstore. Neither collection pipeline draws on COCO, Visual Genome, or Open Images, which underlie our natural-image training sources. Neither draws on our document, chart, or diagram sources either. Both benchmarks are therefore disjoint from our training images at the source level, and no image-ID audit is required.

\section{Budget Matching Against Baselines}
\label{app:budget}

All comparisons in Tab.~\ref{tab:main_results} are budget-matched: each baseline is given a parameter budget comparable to that of the corresponding \textsc{V-Lumen} variant. 
The matched quantity depends on the baseline family. 
For LoRA, we match \emph{trainable} parameters. 
For MoE, whose defining trade-off is between total and per-token activated parameters, we match both \emph{accessible} parameters (all parameters the model can route to) and \emph{activated} parameters per token. 
All \textsc{V-Lumen} variants use the same memory configuration as in Tab.~\ref{tab:main_results}.

\paragraph{LoRA baselines.}
We compare against LoRA under two designs. Here, LoRA($\beta$) denotes a LoRA adapter whose trainable parameter count is $\beta$ times that of the \textsc{V-Lumen} module.

\textit{Design 1: rank-matched.}
The LoRA component is identical on both sides, and the \textsc{V-Lumen} memory table is added on top (Tab.~\ref{tab:budget-lora-rank}). This design isolates the contribution of \textsc{V-Lumen} at a fixed adapter configuration. Since the \textsc{V-Lumen} variant carries more trainable parameters under this design, Design~2 controls for the total budget.
% Design1 - activated params per token 도 측정해서 비교 넣기 

Tab.~\ref{tab:lora_design1} reports accuracy under this design on Qwen3-VL-2B.
Adding \textsc{V-Lumen} to the same adapter leaves the average accuracy unchanged within
the seed spread ($-0.47$pp), with higher accuracy on Frieda and MMMU-Pro, equal
accuracy on ChartMuseum, and lower accuracy on VisRes and MMMU.
 
\begin{table}[h]
\centering
\caption{Accuracy (\%) under LoRA Design 1 (rank-matched) on Qwen3-VL-2B, mean
$\pm$ std over three seeds. Both settings share the same LoRA(0.5) adapter (rank
107; Tab.~\ref{tab:budget-lora-rank}), and \textsc{V-Lumen} is added on top. Avg $\Delta$ is the average gain over
the vanilla backbone (pp). Training schedules follow Tab.~\ref{tab:training_schedule}.}
\label{tab:lora_design1}
\footnotesize
\setlength{\tabcolsep}{4pt}
\begin{tabular}{lccccccc}
\toprule
Setting & ChartMus. & Frieda & VisRes & MMMU & MMMU-Pro & Avg & Avg $\Delta$ \\
\midrule
Vanilla               & 1.69 & 2.27 & 5.00 & 1.63 & 0.30 & 2.18 & -- \\
LoRA(0.5)             & 18.24{\tiny$\pm$0.70} & 5.53{\tiny$\pm$0.69} & 23.63{\tiny$\pm$1.40} & 43.37{\tiny$\pm$0.82} & 28.20{\tiny$\pm$1.21} & 23.79{\tiny$\pm$0.33} & +21.62{\tiny$\pm$0.33} \\
LoRA(0.5) + \textsc{V-Lumen} & 18.24{\tiny$\pm$0.41} & 5.61{\tiny$\pm$0.35} & 21.70{\tiny$\pm$2.39} & 41.91{\tiny$\pm$1.30} & 29.20{\tiny$\pm$0.66} & 23.33{\tiny$\pm$0.02} & +21.15{\tiny$\pm$0.02} \\
\bottomrule
\end{tabular}
\end{table}

\input{tables/budget_lora1_table}

\textit{Design 2: total-budget-matched.}
We increase the LoRA rank until the LoRA-only model matches the combined trainable parameters of LoRA(0.5) + \textsc{V-Lumen} (Tab.~\ref{tab:budget-lora-total}). The resulting baseline is within 0.2\% of the \textsc{V-Lumen} variant in total trainable parameters, so differences between the two cannot be attributed to the size of the trainable budget. Tab.~\ref{tab:main_results} uses Design 2. Design 1 is used for the
InternVL3.5-2B experiments (Appendix~\ref{app:internvl}) and the logit-lens
analysis (\S\ref{sec:analysis}), and serves as an additional reference in the
cost measurements (Appendix~\ref{app:throughput}).

\input{tables/budget_lora2_table}

\paragraph{MoE baselines.}
The MoE-only baseline is configured to match the combined accessible and per-token activated parameters of MoE + \textsc{V-Lumen} (Tab.~\ref{tab:budget-moe}). The \textsc{V-Lumen} variant is slightly below the baseline on both axes ($0.988\times$ accessible, $0.995\times$ activated), so the comparison, if anything, favors the baseline. In both configurations, the MoE module is attached to the same decoder layers as \textsc{V-Lumen} (layers 2 and 8), uses top-1 routing without a shared expert, and is trained with a load-balancing loss weight of $\alpha = 10^{-2}$.
We attach the MoE module to the same layers as \textsc{V-Lumen} so that the two methods differ only in how the added parameters are organized, not in where they are inserted.
Per-side configurations are given in Tab.~\ref{tab:budget-moe-config}.

\input{tables/budget_moe_table}

\section{Inference Cost}
\subsection{End-to-End Latency}
\label{app:throughput}

We measure prefill and decode latency for the six settings of Tab.~\ref{tab:main_results} and for LoRA(0.5), the rank-matched baseline of Design 1 (Appendix~\ref{app:budget}). To avoid run-to-run drift from GPU clock and thermal state, all seven models are loaded in a single process on one RTX 6000 Ada GPU, and the order of settings is shuffled for every sample. We use 60 ChartMuseum samples over two rounds ($n = 120$), bf16, and 10 warm-up runs. Prefill is the time to generate one token; decode is the per-token time obtained from forced 32-token and 1-token generations, $(t_{32} - t_1)/31$.

% \begin{table}[h]
% \centering
% \caption{Prefill and per-token decode latency (ms). $\Delta$ is the change from adding \textsc{V-Lumen} to the corresponding baseline.}
% \label{tab:prefill_decode}
% \small
% \begin{tabular}{lcccc}
% \toprule
% Setting & Prefill & $\Delta$ prefill & Decode / token & $\Delta$ decode \\
% \midrule
% Vanilla                         & 78.7  & --                             & 39.68 & -- \\
% \textsc{V-Lumen}                & 111.7 & +33.0 {\scriptsize(+42\%)} & 39.65 & $-$0.03 \\
% \midrule
% LoRA(0.5)                       & 101.9 & --                             & 60.27 & -- \\
% LoRA(0.5) + \textsc{V-Lumen}    & 132.3 & +30.4 {\scriptsize(+30\%)} & 60.32 & +0.05 \\
% \midrule
% MoE                             & 100.5 & --                             & 39.65 & -- \\
% MoE + \textsc{V-Lumen}          & 124.5 & +24.0 {\scriptsize(+24\%)} & 39.69 & +0.04 \\
% \bottomrule
% \end{tabular}
% \end{table}
\begin{table}[h]
\centering
\caption{Prefill and per-token decode latency (ms). $\Delta$ is the change from
adding \textsc{V-Lumen} to the corresponding budget-matched baseline (Vanilla,
LoRA(1.5), or MoE(1.5)). Relative to the rank-matched LoRA(0.5),
LoRA(0.5) + \textsc{V-Lumen} adds $+25.0$\,ms prefill and $-0.04$\,ms decode.}
\label{tab:latency}
\small
\begin{tabular}{lcccc}
\toprule
Setting & Prefill & $\Delta$ prefill & Decode / token & $\Delta$ decode \\
\midrule
Vanilla               & 66.0  & --           & 31.90 & --      \\
\textsc{V-Lumen}             & 91.3  & +25.3 (+38\%) & 31.93 & +0.03   \\
\midrule
LoRA(0.5)             & 85.2  & --           & 49.07 & --      \\
LoRA(1.5)             & 85.4  & --           & 49.29 & --      \\
LoRA(0.5) + \textsc{V-Lumen} & 110.2 & +24.8 (+29\%) & 49.03 & $-$0.26 \\
\midrule
MoE(1.5)              & 84.2  & --           & 31.96 & --      \\
MoE(0.5) + \textsc{V-Lumen}  & 102.8 & +18.6 (+22\%) & 31.91 & $-$0.05 \\
\bottomrule
\end{tabular}
\end{table}

% \paragraph{Decode.}
% Adding \textsc{V-Lumen} leaves per-token decode latency unchanged in all three pairs ($|\Delta| \le 0.05$\,ms). Memory is retrieved and fused only at visual-token positions. Since visual tokens are processed once during prefill and their memory-augmented states are stored in the KV cache, each decode step, which processes a single new text token, involves no lookup. \textsc{V-Lumen} therefore adds no per-token cost during generation. The higher decode latency of the LoRA settings comes from evaluating an unmerged adapter, a cost that disappears once the adapter is merged; we therefore do not compare throughput against LoRA.
\paragraph{Decode.} Adding \textsc{V-Lumen} leaves per-token decode latency unchanged:
$|\Delta| \le 0.05$\,ms against the vanilla, MoE(1.5), and rank-matched LoRA(0.5)
baselines. The $-0.26$\,ms against LoRA(1.5) reflects its larger rank-321 adapter,
which is 0.22\,ms per token slower than LoRA(0.5), rather than \textsc{V-Lumen}. Memory is
retrieved and fused only at visual-token positions. Since visual tokens are
processed once during prefill and their memory-augmented states are stored in the
KV cache, each decode step, which processes a single new text token, involves no
lookup. \textsc{V-Lumen} therefore adds no per-token cost during generation. The higher
decode latency of the LoRA settings comes from evaluating an unmerged adapter, a
cost that disappears once the adapter is merged; we therefore do not compare
throughput against LoRA.

\paragraph{Prefill.}
\label{app:throughput:prefill}
% \textsc{V-Lumen} adds 24--33\,ms of one-off prefill latency. Tab.~\ref{tab:prefill_breakdown} decomposes this overhead; the component sums match the measured increases (32.5 vs.\ 33.0\,ms and 30.3 vs.\ 30.4\,ms). The \textsc{V-Lumen} cost is about 30\,ms in every setting. The smaller increase in the MoE pair arises because the MoE module of the budget-matched MoE + \textsc{V-Lumen} variant is smaller (21 experts of width 168 versus 31 of width 344), which reduces MoE time by about 6\,ms. Most of the overhead ($\sim$75\%) is key construction, which our current implementation computes with a Python loop on the CPU; the table read itself takes about 2.3\,ms and, as shown in Tab.~\ref{tab:retrieval_latency}, does not grow with table size. Key construction consists of integer hashing and indexing over the visual-token grid, which we expect to be reducible by vectorizing it on the GPU.
\textsc{V-Lumen} adds 19--25\,ms of one-off prefill latency.
Tab.~\ref{tab:prefill_breakdown} decomposes this overhead; the component sums match
the measured increases (24.6 vs.\ 25.3\,ms over the vanilla backbone, and 23.7 vs.\
25.0\,ms over LoRA(0.5), which shares the same adapter). The \textsc{V-Lumen} cost is
about 23--25\,ms in every setting. The smaller increase in the MoE pair arises
because the MoE module of the budget-matched MoE(0.5) + \textsc{V-Lumen} variant is
smaller (21 experts of width 168 versus 31 of width 344), which reduces MoE time
by about 5\,ms (17.0 $\rightarrow$ 12.1\,ms). Most of the overhead (about three
quarters) is key construction, which our current implementation computes with a
Python loop on the CPU; the table read itself takes about 1.8\,ms and, as shown in
Tab.~\ref{tab:retrieval_latency}, does not grow with table size. Key construction
consists of integer hashing and indexing over the visual-token grid, which we
expect to be reducible by vectorizing it on the GPU.

\begin{table}[h]
\centering
\caption{Breakdown of \textsc{V-Lumen} prefill overhead (ms per sample, summed over both
fusion layers).}
\label{tab:prefill_breakdown}
\footnotesize
\begin{tabular}{lcccc}
\toprule
Component & \textsc{V-Lumen} & LoRA(0.5) + \textsc{V-Lumen} & MoE(0.5) + \textsc{V-Lumen} & MoE(1.5) \\
\midrule
Quantizer (patch $\rightarrow$ code)  & 0.3  & 0.3  & 0.3  & -- \\
TCH bins (PRH / DRH)                  & 1.1  & 1.1  & 1.1  & -- \\
Key construction                      & 18.2 & 17.3 & 16.6 & -- \\
Gather (table read)                   & 1.8  & 1.8  & 1.8  & -- \\
Fusion (gate, value projection, norm) & 3.2  & 3.2  & 3.1  & -- \\
\midrule
\textsc{V-Lumen} total                       & 24.6 & 23.7 & 22.8 & -- \\
MoE module                            & --   & --   & 12.1 & 17.0 \\
\bottomrule
\end{tabular}
\end{table}
% \paragraph{End-to-end overhead.}
% Because the overhead is incurred once per input, its relative cost shrinks with answer length. At the average answer lengths observed in our evaluation (about 4.5 tokens for the LoRA settings and 9 tokens for the MoE settings), the measured prefill and decode latencies imply an end-to-end increase of about 7\% for the LoRA pair (379 $\rightarrow$ 404\,ms) and 8\% for the MoE pair (458 $\rightarrow$ 494\,ms).

% Our central efficiency claim is therefore the decoupling of capacity from per-token compute shown in Sec.~\ref{sec:retrieval}, rather than a throughput advantage.

\paragraph{End-to-end overhead.}
For an answer of $n$ generated tokens, end-to-end latency is approximately
\begin{equation}
T(n) = t_{\text{prefill}} + n \cdot t_{\text{decode}}.
\end{equation}

Since \textsc{V-Lumen} leaves $t_{\text{decode}}$ unchanged, its relative overhead is approximately $\Delta t_{\text{prefill}} / T(n)$: the one-off prefill cost is amortized over the generated tokens and shrinks as answers get longer. 
For example, with a 7-token answer, close to the average answer length in our evaluation, Tab.~\ref{tab:latency} gives an end-to-end increase of about 5--6\% for both the LoRA pair (LoRA(1.5) $\rightarrow$ LoRA(0.5) + \textsc{V-Lumen}, 430 $\rightarrow$ 453\,ms) and the MoE pair (308 $\rightarrow$ 326\,ms).
% For example, with a 7-token answer, close to the average answer length in our evaluation, Tab.~\ref{tab:prefill_decode} gives an end-to-end increase of about 6\% for both the LoRA pair (524 $\rightarrow$ 555\,ms) and the MoE pair (378 $\rightarrow$ 402\,ms).

\subsection{Memory Overhead}
% Tab.~\ref{tab:memory_overhead} reports GPU memory after loading each model and the peak during inference (same checkpoints and hardware as Tab.~\ref{tab:latency}). When \textsc{V-Lumen} is added on top of the vanilla backbone or LoRA(0.5), GPU memory increases by 0.35\,GB. Almost all of this is the memory table: our implementation keeps all trainable modules, including the memory table, LoRA adapters, and MoE experts, in fp32 on top of the bf16 backbone, so the table's 83,952,896 parameters occupy 0.34\,GB. In the MoE pair, the only pair in Tab.~\ref{tab:memory_overhead} matched in total parameter budget (Appendix~\ref{app:budget}), the increase is 0.02\,GB (4.781 $\rightarrow$ 4.801\,GB, with the same increase at peak). This is because the MoE component of MoE + \textsc{V-Lumen} is smaller than the MoE-only baseline, which offsets nearly all of the table's footprint; at a matched budget, \textsc{V-Lumen} reallocates memory rather than adding to it. In our current implementation the memory table resides on the GPU; offloading it to host memory is left to future work.
Tab.~\ref{tab:memory_overhead} reports GPU memory after loading each model and the peak during inference (same checkpoints and hardware as  Tab.~\ref{tab:latency}). When \textsc{V-Lumen} is added on top of the vanilla backbone or LoRA(0.5), GPU memory increases by 0.35\,GB. Almost all of this is the memory table: our implementation keeps all trainable modules, including the memory table, LoRA adapters, and MoE experts, in fp32 on top of the bf16 backbone, so the table's 83,952,896 parameters occupy 0.34\,GB. At a matched total budget, the increase nearly vanishes: 0.003\,GB for the LoRA pair (LoRA(1.5) $\rightarrow$ LoRA(0.5) + \textsc{V-Lumen}, 4.772 $\rightarrow$ 4.775\,GB; 0.001\,GB at peak) and 0.02\,GB for the MoE pair (4.781 $\rightarrow$ 4.801\,GB, with the same increase at peak). In both cases, the adapter component of the +\textsc{V-Lumen} variant is smaller than the adapter-only baseline (LoRA rank 107 vs.\ 321; 21 experts of width 168 vs.\ 31 of width 344), which offsets nearly all of the table's footprint; at a matched budget, \textsc{V-Lumen} reallocates memory rather than adding to it. In our current implementation, the memory table resides on the GPU; offloading it to host memory is left to future work.

\begin{table}[h]
\centering
\caption{GPU memory overhead (GB). Memory-table parameters are summed over both fusion layers.}
\label{tab:memory_overhead}
\small
\begin{tabular}{lrcc}
\toprule
Setting & Memory-table params & After load & Peak inference \\
\midrule
Vanilla                       & --                               & 4.256 & 4.311 \\
\textsc{V-Lumen}              & 83,952,896 {\scriptsize(335.8 MB, fp32)} & 4.603 & 4.658 \\
\midrule
LoRA(0.5)                     & --                               & 4.428 & 4.483 \\
LoRA(1.5)                     & --                               & 4.772 & 4.829 \\
LoRA(0.5) + \textsc{V-Lumen}  & 83,952,896 {\scriptsize(335.8 MB, fp32)} & 4.775 & 4.830 \\
\midrule
MoE                           & --                               & 4.781 & 4.836 \\
MoE + \textsc{V-Lumen}        & 83,952,896 {\scriptsize(335.8 MB, fp32)} & 4.801 & 4.856 \\
\bottomrule
\end{tabular}
\end{table}

\section{Additional Experiments}
\subsection{Choice of Visual Codebook}
\textsc{V-Lumen} addresses its memory table through a frozen discretizer, so a natural question is whether the gains depend on the specific codebook of LaVIT~\cite{jin2024lavit}. To test this, we replace LaVIT with SEED~\cite{ge2024making}, an independently pretrained visual tokenizer, and keep everything else identical to the \textsc{V-Lumen}-only setting of Tab.~\ref{tab:main_results}: the backbone, fusion layers (2, 8), memory configuration, training recipe, and data scale. Only the codebook changes (16,384 codes for LaVIT vs.\ 8,192 for SEED); because the discretizer is frozen, both settings have the same number of trainable \textsc{V-Lumen} parameters (86.1M).

\begin{table}[h]
\centering
\caption{\textsc{V-Lumen}-only accuracy (\%) with different frozen codebooks, mean $\pm$ std over three seeds. Avg $\Delta$ is the average gain over the frozen backbone (pp). The LaVIT row is the \textsc{V-Lumen}-only setting of Tab.~\ref{tab:main_results}.}
\label{tab:codebook}
\footnotesize
\setlength{\tabcolsep}{3pt}
\begin{tabular}{lccccc >{\columncolor{gray!10}}c}
\toprule
Codebook & ChartMus. & Frieda & VisRes & MMMU & MMMU-Pro & Avg $\Delta$ \\
\midrule
Vanilla & 1.69 & 2.27 & 5.00 & 1.63 & 0.30 & -- \\
\midrule
LaVIT (16,384 codes)
  & 16.70{\tiny$\pm$0.79} & 4.77{\tiny$\pm$0.23} & 24.50{\tiny$\pm$2.07} & 33.38{\tiny$\pm$0.95} & 20.50{\tiny$\pm$0.82} & +17.79{\tiny$\pm$0.47} \\
SEED (8,192 codes)
  & 16.33{\tiny$\pm$0.52} & 5.83{\tiny$\pm$0.66} & 27.17{\tiny$\pm$1.01} & 34.00{\tiny$\pm$2.63} & 20.10{\tiny$\pm$0.44} & +18.51{\tiny$\pm$1.00} \\
\bottomrule
\end{tabular}
\end{table}

With SEED, \textsc{V-Lumen} improves over the frozen backbone by $+18.51 \pm 1.00$pp on average, $0.72$pp above LaVIT and within the seed spread. SEED is slightly higher on Frieda, VisRes, and MMMU and slightly lower on ChartMuseum and MMMU-Pro, despite using a codebook half the size of LaVIT's. These results indicate that the gains of \textsc{V-Lumen} are not tied to a particular codebook and transfer to an independently pretrained discretizer.

\subsection{Hyperparameter Sensitivity}
\label{app:sensitivity}

\paragraph{Hashing constants.}
The salts, prime multipliers, and rolling-hash coefficients used in \textsc{V-Lumen} (Appendix~\ref{app:hash_constants}) are not tuned hyperparameters. They only serve to give different addressing components (coordinates, text-conditioned bins, and $n$-gram orders) distinct, non-colliding contributions to the hash key, following standard hashing practice. Any coprime primes of similar magnitude behave the same way, so we do not tune them per benchmark.

\paragraph{Single-axis sensitivity.}
We perturb one design choice at a time from the default configuration (fusion layers (2, 8), $K = 8$ text-conditioned bins, 8 memory-table heads, and $n$-gram orders $\{2, 3\}$), keeping all other settings fixed, and report the average gain over the frozen backbone on the five benchmarks of Tab.~\ref{tab:main_results} (Tab.~\ref{tab:sensitivity}). 

Every perturbation yields a gain between $+17.20$ and $+18.64$pp, within $-0.6$ to $+0.9$pp of the default ($+17.79 \pm 0.47$pp) and thus within about one to two seed standard deviations. The benefit of \textsc{V-Lumen} therefore does not hinge on a particular hyperparameter setting, and the default is not uniquely optimal. Moving the fusion layers to (4, 10) is the only perturbation that falls below the default ($-0.59$pp, mostly on MMMU), consistent with the fusion-layer ablation in Appendix~\ref{app:layer_ablation}.

\begin{table}[h]
\centering
\caption{Single-axis sensitivity of \textsc{V-Lumen}-only. Avg $\Delta$ is the average gain over the frozen backbone across the five benchmarks (pp, mean $\pm$ std over three seeds). Changing the number of heads keeps the total memory dimension fixed (512), so $d_{\text{head}}$ becomes 64 with 4 heads and 16 with 16 heads while the table size stays about 86M parameters; this differs from the width axis in Appendix~\ref{app:capacity}, where the total memory dimension changes. Adding a third $n$-gram order (n=4) keeps $d_{\text{head}} = 32$ and adds 16 memory cells.}
\label{tab:sensitivity}
\small
\begin{tabular}{ll >{\columncolor{gray!10}}c}
\toprule
Axis & Setting & Avg $\Delta$ \\
\midrule
--- (default)   & layers (2, 8), $K{=}8$, 8 heads, orders $\{2,3\}$ & +17.79{\scriptsize$\pm$0.47} \\
\midrule
Fusion layers   & (2, 8) $\rightarrow$ (4, 10)                      & +17.20{\scriptsize$\pm$0.35} \\
Bins $K$        & 8 $\rightarrow$ 4                                 & +18.33{\scriptsize$\pm$0.57} \\
                & 8 $\rightarrow$ 16                                & +17.81{\scriptsize$\pm$0.27} \\
Memory heads    & 8 $\rightarrow$ 4                                 & +18.64{\scriptsize$\pm$0.77} \\
                & 8 $\rightarrow$ 16                                & +17.99{\scriptsize$\pm$0.83} \\
$n$-gram orders & $\{2,3\}$ $\rightarrow$ $\{2,3,4\}$               & +17.81{\scriptsize$\pm$0.15} \\
\bottomrule
\end{tabular}
\end{table}

\paragraph{Transfer beyond the training curriculum.}
The same default configuration transfers without modification to three benchmarks outside the training curriculum in a fully zero-shot setting (Tab.~\ref{tab:zeroshot}), suggesting that it is not overfit to the five benchmarks used for development.

\subsection{Generalization to Another Backbone: InternVL3.5-2B}
\label{app:internvl}

To test whether \textsc{V-Lumen} depends on the Qwen3-VL visual pipeline, we repeat the experiments of Tab.~\ref{tab:main_results} on InternVL3.5-2B \cite{wang2025internvl3}, which uses a different vision encoder (InternViT-300M) and a different visual-token pipeline (pixel-shuffle compression) from Qwen3-VL-2B. We use the same recipe as the main experiments: memory width $d_{\text{head}} = 32$, fusion layers (2, 8), text-conditioned hashing with PRH + DRH, and the same training schedules, with $k = 0.5$ (LoRA($k$) corresponds to rank 78 on this backbone). The MoE pair is matched in total parameter budget as in Tab.~\ref{tab:main_results}, whereas the LoRA pair is rank-matched (Design 1 in Appendix~\ref{app:budget}): \textsc{V-Lumen} is added on top of the same LoRA($k$) adapter. The only adaptation is the visual grid: InternVL3.5-2B produces $16 \times 16$ image tokens per image, so spatial aggregation and text-conditioned hashing operate on a $16 \times 16$ grid. Results are mean $\pm$ std over three seeds (Tab.~\ref{tab:internvl}).

\begin{table}[h]
\centering
\caption{Accuracy (\%) on InternVL3.5-2B, mean $\pm$ std over three seeds. Avg $\Delta$ is the average gain over the frozen backbone (pp). The MoE pair is matched in total parameter budget; the LoRA pair is rank-matched (LoRA(0.5) with and without \textsc{V-Lumen}; Design 1 in Appendix~\ref{app:budget}).}
\label{tab:internvl}
\footnotesize
\setlength{\tabcolsep}{3pt}
\begin{tabular}{lcccccc}
\toprule
Setting & ChartMus. & Frieda & VisRes & MMMU & MMMU-Pro & Avg $\Delta$ \\
\midrule
Vanilla & 5.41 & 1.36 & 0.00 & 2.26 & 1.30 & -- \\
\midrule
\textsc{V-Lumen}-only % TODO: update to 3 seeds and remove dagger
  & 14.49{\tiny$\pm$0.86} & 4.62{\tiny$\pm$0.35} & 22.27{\tiny$\pm$0.06} & 29.11{\tiny$\pm$4.79} & 14.33{\tiny$\pm$3.95} & +14.90{\tiny$\pm$1.82} \\
\midrule
LoRA(0.5)
  & 18.62{\tiny$\pm$0.28} & 6.59{\tiny$\pm$0.23} & 21.40{\tiny$\pm$0.61} & 45.17{\tiny$\pm$0.88} & 27.57{\tiny$\pm$0.65} & +21.80{\tiny$\pm$0.28} \\
LoRA(0.5)+ \textsc{V-Lumen}
  & 18.32{\tiny$\pm$0.17} & 5.23{\tiny$\pm$0.39} & 21.80{\tiny$\pm$0.72} & 44.63{\tiny$\pm$2.07} & 26.60{\tiny$\pm$0.87} & +21.25{\tiny$\pm$0.49} \\
\midrule
MoE(1.5)
  & 15.47{\tiny$\pm$0.36} & 6.21{\tiny$\pm$1.05} & 22.53{\tiny$\pm$0.40} & 34.92{\tiny$\pm$1.05} & 18.73{\tiny$\pm$0.81} & +17.51{\tiny$\pm$0.25} \\
MoE(0.5)+ \textsc{V-Lumen}
  & 16.33{\tiny$\pm$0.92} & 5.68{\tiny$\pm$0.99} & 21.70{\tiny$\pm$0.75} & 33.29{\tiny$\pm$0.96} & 19.27{\tiny$\pm$1.76} & +17.19{\tiny$\pm$0.36} \\
\bottomrule
\end{tabular}
\end{table}

The results follow the same pattern as on Qwen3-VL-2B. Used alone, \textsc{V-Lumen} improves the frozen backbone by $+14.90$pp on average, so the mechanism is not specific to the Qwen3-VL visual pipeline. On this backbone, its standalone gain is below that of LoRA-only and MoE-only, although the vanilla-based setting uses a longer training schedule and is not directly comparable to the adapter-based settings (Appendix~\ref{app:training}). In the budget-matched MoE pair, MoE + \textsc{V-Lumen} performs close to MoE-only ($-0.32$pp, within the seed spread), so reallocating part of the budget from expert parameters to \textsc{V-Lumen} preserves performance on a second backbone with a different visual-token pipeline. In the rank-matched LoRA pair, adding \textsc{V-Lumen} on top of the same adapter changes the average by $-0.55$pp, which is comparable to the seed spread and consistent with Design 1 on Qwen3-VL-2B (Tab.~\ref{tab:lora_design1}).

\subsection{Component Ablation and Robustness to Code Changes}
\label{app:ablation}

Tab.~\ref{tab:ablation} ablates the hashing components of \textsc{V-Lumen} (\texttt{SAH}, \texttt{PRH}, and \texttt{DRH}) on the frozen vanilla backbone, where only the memory and fusion parameters are trained. The \textit{rolling hash only} variant replaces spatial aggregation with sequential 2-/3-gram rolling hashing over the visual codes. All results are mean $\pm$ std over three seeds.

\paragraph{Accuracy without perturbation.}
All variants reach a similar average gain over the frozen backbone, from $+17.79$ to $+18.67$pp, and the differences fall within the seed spread. Without perturbation, no single component changes accuracy, and rolling hashing alone already captures most of the gain.

\paragraph{Stress test: re-drawn projection.}
\textsc{V-Lumen} addresses its memory through discrete visual codes, each obtained as $\arg\max_{v} \langle \mathrm{normalize}(W_{\text{proj}}\, x), \bar{c}_v \rangle$, where $x$ is the visual embedding at the LLM input, $W_{\text{proj}}$ is a frozen random projection, and $\bar{c}_v$ are the frozen codebook entries. When the codes change, the memory reads rows that were trained for other codes. To test how each key design copes with this, we discretize a fixed random subset of image patches with a re-drawn frozen projection at evaluation time only: half of the patches in the 50\% setting and all of them in the 100\% setting. Here the components differ clearly. Rolling hash only is the most fragile, losing $4.67$pp at 50\% and $14.69$pp at 100\%, with the largest variance. Adding \texttt{SAH} sharply reduces the loss at 100\% ($-4.25$pp), and adding either relevance-hashing variant alone performs similarly. The full model, \texttt{SAH} + \texttt{PRH} + \texttt{DRH}, shows the smallest drop at both levels ($-1.61$pp and $-2.17$pp), retaining a $+15.62$pp gain even when every code changes. The ordering, rolling hash only $<$ \texttt{SAH} $\approx$ single relevance hashing $<$ full model, holds at both levels.

\paragraph{Discussion.}
This difference between key designs appears only under the stress test. Under common image degradation such as JPEG compression and downscaling, which also change most of the visual codes, all variants, including rolling hash only, remain robust, with accuracy changes below 1pp. A re-drawn projection is harsher because it assigns new codes at random, including to informative patches. Spatial aggregation and text-conditioned hashing may therefore matter in settings where codes shift more systematically than under natural degradation, for example when a trained memory is reused after the vision encoder or connector is updated, or under small adversarial perturbations of the input; we do not evaluate these settings here.

% Two simple explanations for why the full design degrades less than rolling hash only under the stress test can be ruled out. First, its addresses do not survive the perturbation better: when all codes change, every memory address changes for all key designs. Second, it does not rely less on the memory: the average fusion-gate activation is similar across designs (0.61--0.67). We hypothesize that it comes from \emph{row sharing}: spatial aggregation blends each code with its row- and column-wise context before hashing, and text-conditioned hashing further salts the key with the query, so each memory row is reached by many codes and contexts during training and learns more generic content. Rolling-hash keys, in contrast, map nearly one-to-one to code $n$-grams, so reading a wrong row injects noise. We have not verified this hypothesis, and the stress test uses a single draw of the perturbed projection.
We hypothesize that the full design degrades less under the stress test because of \emph{row sharing}, rather than because its addresses survive the perturbation or because it relies less on the memory. Spatial aggregation blends each code with its row- and column-wise context before hashing, and text-conditioned hashing further salts the key with the query, so each memory row is reached by many codes and contexts during training and learns more generic content. Reading a row trained for a different code therefore costs little. Rolling-hash keys, in contrast, map nearly one-to-one to code $n$-grams, so their rows are specific to particular code sequences and reading a wrong row injects noise. The two alternative explanations do not hold: when all codes change, every memory address changes for all key designs, and the average fusion-gate activation is similar across designs (0.61--0.67). We have not verified this hypothesis directly, and the stress test uses a single draw of the perturbed projection.

\begin{table}[t]
\centering
\caption{Component ablation of \textsc{V-Lumen} on the frozen vanilla backbone. Values are the average gain over the frozen backbone across five benchmarks (pp, mean $\pm$ std over three seeds). \emph{Perturbed} columns discretize 50\% or 100\% of the image patches with a re-drawn projection at evaluation time; the drop from the unperturbed setting is shown in parentheses, and the smallest drop at each level is \textbf{bold}.}
\label{tab:ablation}
\small
\setlength{\tabcolsep}{4pt}
\begin{tabular}{l c cc}
\toprule
 & & \multicolumn{2}{c}{Perturbed patches} \\
\cmidrule(lr){3-4}
Variant & Unperturbed & 50\% & 100\% \\
\midrule
\textit{Rolling hash only}
  & +18.67{\scriptsize$\pm$0.70} & +14.01{\scriptsize$\pm$1.38} {\scriptsize($-$4.67)} & \phantom{0}+3.98{\scriptsize$\pm$0.21} {\scriptsize($-$14.69)} \\
\texttt{SAH}
  & +18.19{\scriptsize$\pm$0.90} & +15.09{\scriptsize$\pm$0.29} {\scriptsize($-$3.10)} & +13.94{\scriptsize$\pm$0.68} {\scriptsize($-$4.25)} \\
\texttt{SAH} + \texttt{PRH}
  & +18.51{\scriptsize$\pm$0.02} & +15.03{\scriptsize$\pm$1.36} {\scriptsize($-$3.48)} & +13.83{\scriptsize$\pm$0.96} {\scriptsize($-$4.68)} \\
\texttt{SAH} + \texttt{DRH}
  & +18.32{\scriptsize$\pm$0.41} & +15.14{\scriptsize$\pm$0.44} {\scriptsize($-$3.17)} & +14.55{\scriptsize$\pm$1.00} {\scriptsize($-$3.77)} \\
\texttt{SAH} + \texttt{PRH} + \texttt{DRH}
  & +17.79{\scriptsize$\pm$0.47} & +16.19{\scriptsize$\pm$0.31} {\scriptsize\textbf{($-$1.61)}} & +15.62{\scriptsize$\pm$0.11} {\scriptsize\textbf{($-$2.17)}} \\
\bottomrule
\end{tabular}
\end{table}

\subsection{Fusion-Layer Ablation}
\label{app:layer_ablation}

Tab.~\ref{tab:layer_ablation} varies the two decoder layers at which
\textsc{V-Lumen} fuses the retrieved memory, in the
\textsc{V-Lumen}-only setting on the frozen backbone.
All other settings follow Tab.~\ref{tab:main_results},
and results are reported as mean $\pm$ std over three seeds.

\begin{table}[h]
\centering
\caption{Fusion-layer ablation of \textsc{V-Lumen}-only on the frozen
backbone (accuracy \%, mean $\pm$ std over three seeds).
Avg $\Delta$ denotes the average gain over the frozen backbone (pp).
The default configuration is (2, 8).}
\label{tab:layer_ablation}
\footnotesize
\setlength{\tabcolsep}{3pt}
\begin{tabular}{lccccc >{\columncolor{gray!10}}c}
\toprule
Fusion layers & ChartMus. & Frieda & VisRes & MMMU & MMMU-Pro & Avg $\Delta$ \\
\midrule
Vanilla
& 1.69 & 2.27 & 5.00 & 1.63 & 0.30 & -- \\
\midrule
(2, 6)
& 15.92{\tiny$\pm$0.34}
& 5.53{\tiny$\pm$1.07}
& 25.83{\tiny$\pm$1.45}
& 34.42{\tiny$\pm$1.43}
& 19.10{\tiny$\pm$0.92}
& +17.98{\tiny$\pm$0.28} \\
(2, 8)
& 16.70{\tiny$\pm$0.79}
& 4.77{\tiny$\pm$0.23}
& 24.50{\tiny$\pm$2.07}
& 33.38{\tiny$\pm$0.95}
& 20.50{\tiny$\pm$0.82}
& +17.79{\tiny$\pm$0.47} \\
(2, 10)
& 16.07{\tiny$\pm$0.33}
& 5.30{\tiny$\pm$0.13}
& 26.13{\tiny$\pm$1.18}
& 27.44{\tiny$\pm$0.51}
& 16.53{\tiny$\pm$1.72}
& +16.12{\tiny$\pm$0.31} \\
(2, 15)
& 16.03{\tiny$\pm$0.13}
& 5.30{\tiny$\pm$0.47}
& 21.67{\tiny$\pm$1.52}
& 29.57{\tiny$\pm$0.92}
& 17.67{\tiny$\pm$0.91}
& +15.87{\tiny$\pm$0.14} \\
(3, 15)
& 14.86{\tiny$\pm$0.79}
& 4.92{\tiny$\pm$0.47}
& 23.57{\tiny$\pm$0.51}
& 26.68{\tiny$\pm$1.70}
& 16.27{\tiny$\pm$1.48}
& +15.08{\tiny$\pm$0.56} \\
\bottomrule
\end{tabular}
\end{table}

% % --- Version B: if some configurations are clearly lower ---
% \textcolor{blue}{[Dummy Writing] All configurations yield a gain over the frozen backbone, but fusing at later layers reduces it: [TODO: describe trend, e.g., configurations whose second fusion layer is at 15 reach only +X pp]. Fusing memory early enough leaves the remaining layers of the frozen backbone room to integrate the retrieved information. We use (2, 8) as the default throughout the paper.}
Fusing at layers (2, 6) and (2, 8) yields comparable gains over the frozen backbone (+17.98 $\pm$ 0.28 vs. +17.79 $\pm$ 0.47 pp), with the difference falling within seed variance. Moving the second fusion layer deeper generally reduces the gain, reaching +16.12 pp at (2, 10) and +15.87 pp at (2, 15). Starting fusion at layer 3 and ending at layer 15 yields the smallest gain (+15.08 $\pm$ 0.56 pp). These results support fusing at early layers; we use (2, 8) as the default throughout the paper.

% \subsection{Capacity Scaling}
\section{Capacity Scaling: Memory Width and Depth}
\label{app:capacity}

We scale the memory of \textsc{V-Lumen}-only along two axes while keeping all other settings at the default configuration of Tab.~\ref{tab:main_results} (Tab.~\ref{tab:capacity}). Along \emph{depth}, we vary the number of rows per table cell from $0.5\times$ to $8\times$ the default (81,920). This grows the trainable \textsc{V-Lumen} parameters from 44.2M to 673.4M but leaves the computation per token unchanged, since each visual token still reads one row from each of the 32 cells. Along \emph{width}, we vary the per-head memory dimension $d_{\text{head}} \in \{8, 16, 32, 64\}$, which widens every retrieved vector and the fusion projections that read it, so both the accessible and the per-token activated parameters grow (21.6M to 172.2M). Results are mean $\pm$ std over three seeds.

\begin{table}[h]
\centering
\caption{Capacity scaling of \textsc{V-Lumen}-only on the frozen backbone. \emph{Read / token} is the amount of memory read per visual token relative to the default. Avg $\Delta$ is the average gain over the frozen backbone across five benchmarks (pp, mean $\pm$ std over three seeds). $^\ast$Default configuration (Tab.~\ref{tab:main_results}).}
\label{tab:capacity}
\small
\setlength{\tabcolsep}{5pt}
\begin{tabular}{llcc >{\columncolor{gray!10}}c}
\toprule
Axis & Setting & \textsc{V-Lumen} params & Read / token & Avg $\Delta$ \\
\midrule
Depth & $0.5\times$ rows (40,960)        & 44.2M  & $1\times$    & +18.14{\scriptsize$\pm$0.95} \\
      & $1\times$ rows (81,920)$^\ast$   & 86.1M  & $1\times$    & +17.79{\scriptsize$\pm$0.47} \\
      & $2\times$ rows (163,840)         & 170.0M & $1\times$    & +18.36{\scriptsize$\pm$0.84} \\
      & $4\times$ rows (327,680)         & 337.8M & $1\times$    & +18.33{\scriptsize$\pm$0.78} \\
      & $8\times$ rows (655,360)         & 673.4M & $1\times$    & +17.79{\scriptsize$\pm$0.46} \\
\midrule
Width & $d_{\text{head}} = 8$            & 21.6M  & $0.25\times$ & +16.44{\scriptsize$\pm$0.46} \\
      & $d_{\text{head}} = 16$           & 43.1M  & $0.5\times$  & +17.68{\scriptsize$\pm$1.60} \\
      & $d_{\text{head}} = 32^\ast$      & 86.1M  & $1\times$    & +17.79{\scriptsize$\pm$0.47} \\
      & $d_{\text{head}} = 64$           & 172.2M & $2\times$    & +18.70{\scriptsize$\pm$0.54} \\
\bottomrule
\end{tabular}
\end{table}

\paragraph{Depth.}
From $0.5\times$ to $8\times$ the default number of rows, a $15\times$ range in parameters (44.2M to 673.4M), the average gain stays between $+17.79$ and $+18.36$pp, within one seed standard deviation of each configuration. Because lookup cost is constant in table size (Tab.~\ref{tab:retrieval_latency}), these larger tables add capacity in storage without adding per-token compute, but in our current setting the additional rows have not yet translated into higher accuracy. Improving how a larger table is filled during training and read at inference is therefore a natural direction for making this axis pay off (\S\ref{sec:future}).

\paragraph{Width.}
Accuracy instead increases with the width of the retrieved vectors: the average gain rises from $+16.44$pp at $d_{\text{head}} = 8$ to $+17.68$, $+17.79$, and $+18.70$pp at 16, 32, and 64. The $+2.3$pp difference between 8 and 64 comes mostly from MMMU ($+4.8$pp) and MMMU-Pro ($+5.2$pp). The same tendency appears at matched parameter counts: with about 43--44M parameters, a wider memory with fewer rows ($d_{\text{head}} = 32$, $0.5\times$ rows) reaches $+18.14$pp versus $+17.68$pp for $d_{\text{head}} = 16$, and with about 170M parameters, $d_{\text{head}} = 64$ reaches $+18.70$pp versus $+18.36$pp for $2\times$ rows, although both differences lie within the seed spread. Since width also increases the memory read per token, this is not a pure storage effect; even at $d_{\text{head}} = 64$, however, the parameters activated per token by \textsc{V-Lumen} remain a small fraction of the backbone's (about 0.1\% at $d_{\text{head}} = 32$; Tab.~\ref{tab:budget-moe}). We use $d_{\text{head}} = 32$ throughout the paper as a balance between accuracy and memory size: doubling the width to 64 adds $0.9$pp at twice the memory parameters.

\section{Logit-Lens Analysis Details}
\label{app:logit_lens}

\begin{figure}[h]
\centering
\includegraphics[width=0.8\linewidth]{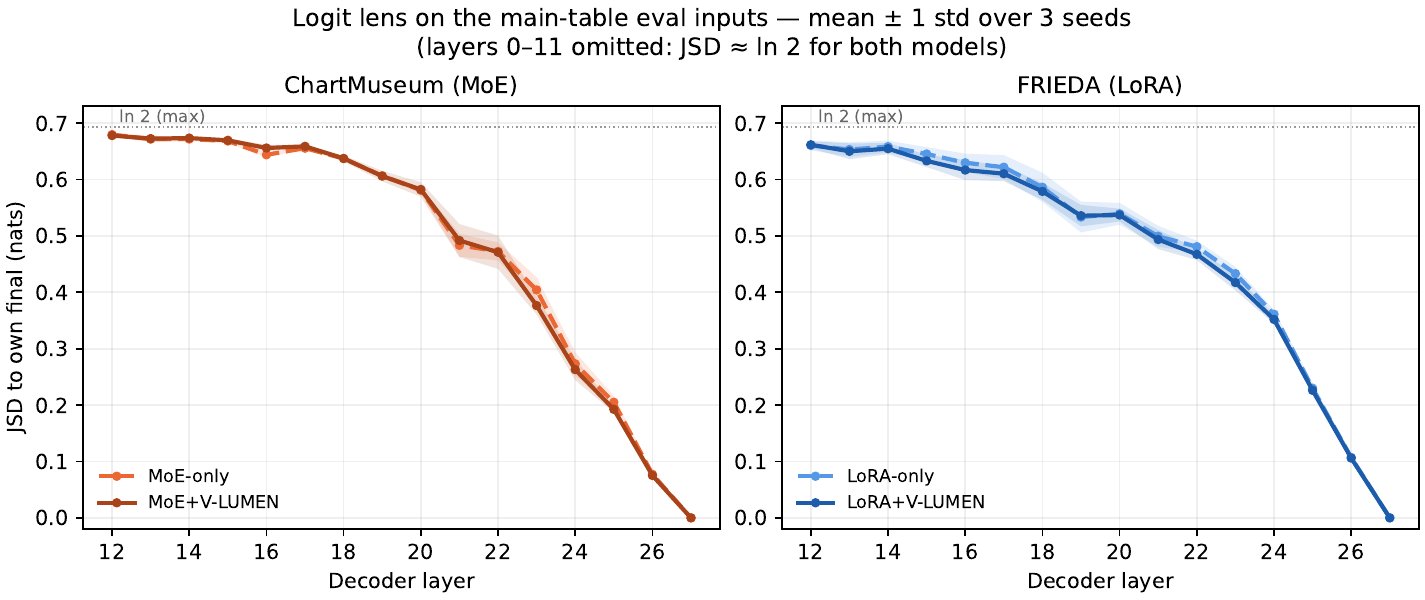}
\caption{Layer-wise $\mathrm{JSD}(p_\ell, p_{\text{final}})$ on the two open-ended probe pairs, ChartMuseum (MoE, left) and Frieda (LoRA, right), comparing each adapter-only model with its +\textsc{V-Lumen} counterpart (MoE: budget-matched; LoRA: rank-matched LoRA(0.5)). Mean $\pm$ 1 std over three seeds.}
\label{fig:logit_lens_all}
\end{figure}

\paragraph{Selection of probe pairs.}
% We fix the selection rule before running the probes: the four (adapter, benchmark) pairs in Tab.~\ref{tab:main_results} where the $+$\textsc{V-Lumen} mean exceeds the adapter-only mean, namely MMMU-Pro (LoRA), VisRes (MoE), FRIEDA (LoRA), and ChartMuseum (MoE). MMMU-Pro and VisRes have single-letter answers and are analyzed with both JSD and gold top-1 accuracy (Fig.~\ref{fig:logit_lens}). FRIEDA and ChartMuseum have open-ended, multi-token answers and are analyzed with JSD only (Fig.~\ref{fig:logit_lens_all}).
We fix the selection rule before running the probes: the four (adapter, benchmark) pairs in which the +\textsc{V-Lumen} mean exceeds the adapter-only mean, using the budget-matched MoE pair of Tab.~\ref{tab:main_results} and the rank-matched LoRA(0.5) pair of Tab.~\ref{tab:lora_design1}, namely MMMU-Pro (LoRA), VisRes (MoE), FRIEDA (LoRA), and ChartMuseum (MoE). MMMU-Pro and VisRes have single-letter answers and are analyzed with both JSD and gold top-1 accuracy (Fig.~\ref{fig:logit_lens}). Frieda and ChartMuseum have open-ended, multi-token answers and are analyzed with JSD only (Fig.~\ref{fig:logit_lens_all}).

\paragraph{Consistency with the main evaluation.}
Inputs and samples are identical to the main evaluation ($n = 900$ for MMMU-Pro and VisRes, 888 for ChartMuseum, and 438 for FRIEDA). For the $+$\textsc{V-Lumen} models, the final-layer top-1 prediction equals the first generated token of the main evaluation for every sample. Text-conditioned hashing requires a single grid of image tokens and is therefore applied only to single-image samples, as in the main evaluation; 7\% of MMMU-Pro and 61\% of FRIEDA samples contain multiple images.

\paragraph{Results on open-ended benchmarks.}
The two open-ended pairs (Fig.~\ref{fig:logit_lens_all}) show the same tendency as the single-letter pairs in Fig.~\ref{fig:logit_lens}. Averaged over layers 19--26, the JSD with \textsc{V-Lumen} is slightly lower on both FRIEDA and ChartMuseum ($-0.006 \pm 0.011$ in each case), and on FRIEDA it is lower around layers 15--17 and 22--23 ($-0.01$), indicating that the model approaches its final prediction somewhat earlier. ChartMuseum also shows a small increase around layer 16 ($+0.012 \pm 0.001$, positive in all three seeds) before the JSD falls below the baseline around layers 23--25; this may reflect a brief perturbation of the intermediate trajectory by the retrieved memory, although its magnitude is small. As with the single-letter pairs, these differences are small relative to seed variance.

%% file: tables/budget_lora1_table.tex
\begin{table}[h]
\centering
\caption{LoRA Design 1 (rank-matched). The LoRA component is shared, and \textsc{V-Lumen} is added on top.}
\label{tab:budget-lora-rank}
\small
\begin{tabular}{lcrrr}
\toprule
Configuration & LoRA rank & LoRA params & \textsc{V-Lumen} params & Total trainable \\
\midrule
LoRA(0.5)                    & 107 & 42,950,656 & --         & 42,950,656  \\
LoRA(0.5) + \textsc{V-Lumen} & 107 & 42,950,656 & 86,132,038 & 129,082,694 \\
\bottomrule
\end{tabular}
\end{table}

%% file: tables/budget_lora2_table.tex
\begin{table}[h]
\centering
\caption{LoRA Design 2 (total-budget-matched). The LoRA-only baseline is scaled to match the combined trainable parameters of LoRA(0.5) + \textsc{V-Lumen}.}
\label{tab:budget-lora-total}
\small
\begin{tabular}{lcrrrc}
\toprule
Configuration & LoRA rank & LoRA params & \textsc{V-Lumen} params & Total trainable & Ratio \\
\midrule
LoRA(0.5) + \textsc{V-Lumen} & 107 & 42,950,656  & 86,132,038 & 129,082,694 & 1.000$\times$ \\
LoRA(1.5)                    & 321 & 128,851,968 & --         & 128,851,968 & 0.998$\times$ \\
\bottomrule
\end{tabular}
\end{table}

%% file: tables/budget_moe_table.tex
\begin{table}[h]
\centering
\caption{MoE budget matching. Accessible parameters include all parameters the model can route to; activated parameters are those used per token.}
\label{tab:budget-moe}
\small
\begin{tabular}{lrr}
\toprule
Configuration & Accessible params & Activated params / token \\
\midrule
MoE only                          & 131,170,304 & 4,358,144 \\
\midrule
MoE + \textsc{V-Lumen}            & 129,574,214 {\scriptsize(0.988$\times$)} & 4,334,592 {\scriptsize(0.995$\times$)} \\
\quad MoE component               & 43,442,176  & 2,154,496 \\
\quad \textsc{V-Lumen} component  & 86,132,038  & 2,180,096 \\
\bottomrule
\end{tabular}
\end{table}

\begin{table}[h]
\centering
\caption{Per-side MoE configurations. Both use top-1 routing at decoder layers 2 and 8 without a shared expert. The \textsc{V-Lumen} memory table consists of 32 cells, one for each combination of fusion layer (2), $n$-gram order (2), and head (8).}
\label{tab:budget-moe-config}
\small
\begin{tabular}{lccc}
\toprule
Configuration & \# experts & Expert hidden dim & \textsc{V-Lumen} memory table \\
\midrule
MoE only              & 31 & 344 & -- \\
MoE + \textsc{V-Lumen} & 21 & 168 & 32 cells $\times$ 81,920 rows $\times$ 32 dims \\
\bottomrule
\end{tabular}
\end{table}

%% file: checklist.tex
\section*{NeurIPS Paper Checklist}

%%% BEGIN INSTRUCTIONS %%%
The checklist is designed to encourage best practices for responsible machine learning research, addressing issues of reproducibility, transparency, research ethics, and societal impact. Do not remove the checklist: {\bf The papers not including the checklist will be desk rejected.} The checklist should follow the references and follow the (optional) supplemental material.  The checklist does NOT count towards the page
limit. 

Please read the checklist guidelines carefully for information on how to answer these questions. For each question in the checklist:
\begin{itemize}
    \item You should answer \answerYes{}, \answerNo{}, or \answerNA{}.
    \item \answerNA{} means either that the question is Not Applicable for that particular paper or the relevant information is Not Available.
    \item Please provide a short (1--2 sentence) justification right after your answer (even for \answerNA). 
   % \item {\bf The papers not including the checklist will be desk rejected.}
\end{itemize}

{\bf The checklist answers are an integral part of your paper submission.} They are visible to the reviewers, area chairs, senior area chairs, and ethics reviewers. You will also be asked to include it (after eventual revisions) with the final version of your paper, and its final version will be published with the paper.

The reviewers of your paper will be asked to use the checklist as one of the factors in their evaluation. While \answerYes{} is generally preferable to \answerNo{}, it is perfectly acceptable to answer \answerNo{} provided a proper justification is given (e.g., error bars are not reported because it would be too computationally expensive'' or ``we were unable to find the license for the dataset we used''). In general, answering \answerNo{} or \answerNA{} is not grounds for rejection. While the questions are phrased in a binary way, we acknowledge that the true answer is often more nuanced, so please just use your best judgment and write a justification to elaborate. All supporting evidence can appear either in the main paper or the supplemental material, provided in appendix. If you answer \answerYes{} to a question, in the justification please point to the section(s) where related material for the question can be found.

IMPORTANT, please:
\begin{itemize}
    \item {\bf Delete this instruction block, but keep the section heading ``NeurIPS Paper Checklist"},
    \item  {\bf Keep the checklist subsection headings, questions/answers and guidelines below.}
    \item {\bf Do not modify the questions and only use the provided macros for your answers}.
\end{itemize}

%%% END INSTRUCTIONS %%%

\begin{enumerate}

\item {\bf Claims}
    \item[] Question: Do the main claims made in the abstract and introduction accurately reflect the paper's contributions and scope?
    \item[] Answer: \answerYes{} % Replace by \answerYes{}, \answerNo{}, or \answerNA{}.
    \item[] Justification: The abstract and introduction state the main contributions of \textsc{V-Lumen}, and the experimental sections evaluate the claimed benefits across frozen-backbone, LoRA-based, and MoE-based settings.
    
    \item[] Guidelines:
    \begin{itemize}
        \item The answer \answerNA{} means that the abstract and introduction do not include the claims made in the paper.
        \item The abstract and/or introduction should clearly state the claims made, including the contributions made in the paper and important assumptions and limitations. A \answerNo{} or \answerNA{} answer to this question will not be perceived well by the reviewers. 
        \item The claims made should match theoretical and experimental results, and reflect how much the results can be expected to generalize to other settings. 
        \item It is fine to include aspirational goals as motivation as long as it is clear that these goals are not attained by the paper. 
    \end{itemize}

\item {\bf Limitations}
    \item[] Question: Does the paper discuss the limitations of the work performed by the authors?
    \item[] Answer: \answerYes{} % Replace by \answerYes{}, \answerNo{}, or \answerNA{}.
    \item[] Justification: The paper discusses its limitations in Appendix~\ref{app:limitations}, including that offloading the memory is a design property we do not measure.
    \item[] Guidelines:
    \begin{itemize}
        \item The answer \answerNA{} means that the paper has no limitation while the answer \answerNo{} means that the paper has limitations, but those are not discussed in the paper. 
        \item The authors are encouraged to create a separate ``Limitations'' section in their paper.
        \item The paper should point out any strong assumptions and how robust the results are to violations of these assumptions (e.g., independence assumptions, noiseless settings, model well-specification, asymptotic approximations only holding locally). The authors should reflect on how these assumptions might be violated in practice and what the implications would be.
        \item The authors should reflect on the scope of the claims made, e.g., if the approach was only tested on a few datasets or with a few runs. In general, empirical results often depend on implicit assumptions, which should be articulated.
        \item The authors should reflect on the factors that influence the performance of the approach. For example, a facial recognition algorithm may perform poorly when image resolution is low or images are taken in low lighting. Or a speech-to-text system might not be used reliably to provide closed captions for online lectures because it fails to handle technical jargon.
        \item The authors should discuss the computational efficiency of the proposed algorithms and how they scale with dataset size.
        \item If applicable, the authors should discuss possible limitations of their approach to address problems of privacy and fairness.
        \item While the authors might fear that complete honesty about limitations might be used by reviewers as grounds for rejection, a worse outcome might be that reviewers discover limitations that aren't acknowledged in the paper. The authors should use their best judgment and recognize that individual actions in favor of transparency play an important role in developing norms that preserve the integrity of the community. Reviewers will be specifically instructed to not penalize honesty concerning limitations.
    \end{itemize}

\item {\bf Theory assumptions and proofs}
    \item[] Question: For each theoretical result, does the paper provide the full set of assumptions and a complete (and correct) proof?
    \item[] Answer: \answerNo{} % Replace by \answerYes{}, \answerNo{}, or \answerNA{}.
    \item[] Justification: The paper does not present formal theoretical results or theorem-proof claims.
    \item[] Guidelines:
    \begin{itemize}
        \item The answer \answerNA{} means that the paper does not include theoretical results. 
        \item All the theorems, formulas, and proofs in the paper should be numbered and cross-referenced.
        \item All assumptions should be clearly stated or referenced in the statement of any theorems.
        \item The proofs can either appear in the main paper or the supplemental material, but if they appear in the supplemental material, the authors are encouraged to provide a short proof sketch to provide intuition. 
        \item Inversely, any informal proof provided in the core of the paper should be complemented by formal proofs provided in appendix or supplemental material.
        \item Theorems and Lemmas that the proof relies upon should be properly referenced. 
    \end{itemize}

    \item {\bf Experimental result reproducibility}
    \item[] Question: Does the paper fully disclose all the information needed to reproduce the main experimental results of the paper to the extent that it affects the main claims and/or conclusions of the paper (regardless of whether the code and data are provided or not)?
    \item[] Answer: \answerYes{} % Replace by \answerYes{}, \answerNo{}, or \answerNA{}.
    \item[] Justification: The paper describes the model architecture, training datasets, evaluation protocol, baselines, training objective, and key hyperparameters needed to reproduce the main experimental results.
    \item[] Guidelines:
    \begin{itemize}
        \item The answer \answerNA{} means that the paper does not include experiments.
        \item If the paper includes experiments, a \answerNo{} answer to this question will not be perceived well by the reviewers: Making the paper reproducible is important, regardless of whether the code and data are provided or not.
        \item If the contribution is a dataset and\slash or model, the authors should describe the steps taken to make their results reproducible or verifiable. 
        \item Depending on the contribution, reproducibility can be accomplished in various ways. For example, if the contribution is a novel architecture, describing the architecture fully might suffice, or if the contribution is a specific model and empirical evaluation, it may be necessary to either make it possible for others to replicate the model with the same dataset, or provide access to the model. In general. releasing code and data is often one good way to accomplish this, but reproducibility can also be provided via detailed instructions for how to replicate the results, access to a hosted model (e.g., in the case of a large language model), releasing of a model checkpoint, or other means that are appropriate to the research performed.
        \item While NeurIPS does not require releasing code, the conference does require all submissions to provide some reasonable avenue for reproducibility, which may depend on the nature of the contribution. For example
        \begin{enumerate}
            \item If the contribution is primarily a new algorithm, the paper should make it clear how to reproduce that algorithm.
            \item If the contribution is primarily a new model architecture, the paper should describe the architecture clearly and fully.
            \item If the contribution is a new model (e.g., a large language model), then there should either be a way to access this model for reproducing the results or a way to reproduce the model (e.g., with an open-source dataset or instructions for how to construct the dataset).
            \item We recognize that reproducibility may be tricky in some cases, in which case authors are welcome to describe the particular way they provide for reproducibility. In the case of closed-source models, it may be that access to the model is limited in some way (e.g., to registered users), but it should be possible for other researchers to have some path to reproducing or verifying the results.
        \end{enumerate}
    \end{itemize}

\item {\bf Open access to data and code}
    \item[] Question: Does the paper provide open access to the data and code, with sufficient instructions to faithfully reproduce the main experimental results, as described in supplemental material?
    \item[] Answer: \answerYes{}
    \item[] Justification: The paper uses public datasets and describes the experimental protocol.
    \item[] Guidelines:
    \begin{itemize}
        \item The answer \answerNA{} means that paper does not include experiments requiring code.
        \item Please see the NeurIPS code and data submission guidelines (\url{https://neurips.cc/public/guides/CodeSubmissionPolicy}) for more details.
        \item While we encourage the release of code and data, we understand that this might not be possible, so \answerNo{} is an acceptable answer. Papers cannot be rejected simply for not including code, unless this is central to the contribution (e.g., for a new open-source benchmark).
        \item The instructions should contain the exact command and environment needed to run to reproduce the results. See the NeurIPS code and data submission guidelines (\url{https://neurips.cc/public/guides/CodeSubmissionPolicy}) for more details.
        \item The authors should provide instructions on data access and preparation, including how to access the raw data, preprocessed data, intermediate data, and generated data, etc.
        \item The authors should provide scripts to reproduce all experimental results for the new proposed method and baselines. If only a subset of experiments are reproducible, they should state which ones are omitted from the script and why.
        \item At submission time, to preserve anonymity, the authors should release anonymized versions (if applicable).
        \item Providing as much information as possible in supplemental material (appended to the paper) is recommended, but including URLs to data and code is permitted.
    \end{itemize}

\item {\bf Experimental setting/details}
    \item[] Question: Does the paper specify all the training and test details (e.g., data splits, hyperparameters, how they were chosen, type of optimizer) necessary to understand the results?
    \item[] Answer: \answerYes{} % Replace by \answerYes{}, \answerNo{}, or \answerNA{}.
    \item[] Justification: The paper specifies the backbone model, training stages, data mixtures, evaluation splits, baselines, optimizer, learning-rate schedule, precision, and key implementation details in the main text and appendix.
    \item[] Guidelines:
    \begin{itemize}
        \item The answer \answerNA{} means that the paper does not include experiments.
        \item The experimental setting should be presented in the core of the paper to a level of detail that is necessary to appreciate the results and make sense of them.
        \item The full details can be provided either with the code, in appendix, or as supplemental material.
    \end{itemize}

\item {\bf Experiment statistical significance}
    \item[] Question: Does the paper report error bars suitably and correctly defined or other appropriate information about the statistical significance of the experiments?
    \item[] Answer: \answerYes{} % Replace by \answerYes{}, \answerNo{}, or \answerNA{}.
    \item[] Justification: Main results, zero-shot results, the second-backbone results, and the ablations report mean $\pm$ standard deviation over three seeds; the layer-wise logit-lens analysis reports mean $\pm$ standard deviation over three seeds.
    \item[] Guidelines:
    \begin{itemize}
        \item The answer \answerNA{} means that the paper does not include experiments.
        \item The authors should answer \answerYes{} if the results are accompanied by error bars, confidence intervals, or statistical significance tests, at least for the experiments that support the main claims of the paper.
        \item The factors of variability that the error bars are capturing should be clearly stated (for example, train/test split, initialization, random drawing of some parameter, or overall run with given experimental conditions).
        \item The method for calculating the error bars should be explained (closed form formula, call to a library function, bootstrap, etc.)
        \item The assumptions made should be given (e.g., Normally distributed errors).
        \item It should be clear whether the error bar is the standard deviation or the standard error of the mean.
        \item It is OK to report 1-sigma error bars, but one should state it. The authors should preferably report a 2-sigma error bar than state that they have a 96\% CI, if the hypothesis of Normality of errors is not verified.
        \item For asymmetric distributions, the authors should be careful not to show in tables or figures symmetric error bars that would yield results that are out of range (e.g., negative error rates).
        \item If error bars are reported in tables or plots, the authors should explain in the text how they were calculated and reference the corresponding figures or tables in the text.
    \end{itemize}

\item {\bf Experiments compute resources}
    \item[] Question: For each experiment, does the paper provide sufficient information on the computer resources (type of compute workers, memory, time of execution) needed to reproduce the experiments?
    \item[] Answer: \answerYes{} % Replace by \answerYes{}, \answerNo{}, or \answerNA{}.
    \item[] Justification: Appendix~\ref{app:compute} reports the GPU type and number used for training, the wall-clock time of each training configuration and of evaluation, and the total compute used for the reported experiments; training schedules and batch sizes are given in Appendix~\ref{app:training}, and latency and GPU memory measurements, including the hardware used, are reported in Appendix~\ref{app:throughput}.
    \item[] Guidelines:
    \begin{itemize}
        \item The answer \answerNA{} means that the paper does not include experiments.
        \item The paper should indicate the type of compute workers CPU or GPU, internal cluster, or cloud provider, including relevant memory and storage.
        \item The paper should provide the amount of compute required for each of the individual experimental runs as well as estimate the total compute. 
        \item The paper should disclose whether the full research project required more compute than the experiments reported in the paper (e.g., preliminary or failed experiments that didn't make it into the paper). 
    \end{itemize}
    
\item {\bf Code of ethics}
    \item[] Question: Does the research conducted in the paper conform, in every respect, with the NeurIPS Code of Ethics \url{https://neurips.cc/public/EthicsGuidelines}?
    \item[] Answer: \answerYes{}% Replace by \answerYes{}, \answerNo{}, or \answerNA{}.
    \item[] Justification: The work uses public benchmarks and pretrained models for methodological research, and we have reviewed the NeurIPS Code of Ethics and believe the research conforms to it.
    \item[] Guidelines:
    \begin{itemize}
        \item The answer \answerNA{} means that the authors have not reviewed the NeurIPS Code of Ethics.
        \item If the authors answer \answerNo, they should explain the special circumstances that require a deviation from the Code of Ethics.
        \item The authors should make sure to preserve anonymity (e.g., if there is a special consideration due to laws or regulations in their jurisdiction).
    \end{itemize}

\item {\bf Broader impacts}
    \item[] Question: Does the paper discuss both potential positive societal impacts and negative societal impacts of the work performed?
    \item[] Answer: \answerYes{} % Replace by \answerYes{}, \answerNo{}, or \answerNA{}.
    \item[] Justification: The paper discusses that improving VLM reasoning can benefit document, chart, and multimodal understanding applications, while also acknowledging potential risks from more capable VLMs such as incorrect or overconfident predictions in downstream use.
    \item[] Guidelines:
    \begin{itemize}
        \item The answer \answerNA{} means that there is no societal impact of the work performed.
        \item If the authors answer \answerNA{} or \answerNo, they should explain why their work has no societal impact or why the paper does not address societal impact.
        \item Examples of negative societal impacts include potential malicious or unintended uses (e.g., disinformation, generating fake profiles, surveillance), fairness considerations (e.g., deployment of technologies that could make decisions that unfairly impact specific groups), privacy considerations, and security considerations.
        \item The conference expects that many papers will be foundational research and not tied to particular applications, let alone deployments. However, if there is a direct path to any negative applications, the authors should point it out. For example, it is legitimate to point out that an improvement in the quality of generative models could be used to generate Deepfakes for disinformation. On the other hand, it is not needed to point out that a generic algorithm for optimizing neural networks could enable people to train models that generate Deepfakes faster.
        \item The authors should consider possible harms that could arise when the technology is being used as intended and functioning correctly, harms that could arise when the technology is being used as intended but gives incorrect results, and harms following from (intentional or unintentional) misuse of the technology.
        \item If there are negative societal impacts, the authors could also discuss possible mitigation strategies (e.g., gated release of models, providing defenses in addition to attacks, mechanisms for monitoring misuse, mechanisms to monitor how a system learns from feedback over time, improving the efficiency and accessibility of ML).
    \end{itemize}
    
\item {\bf Safeguards}
    \item[] Question: Does the paper describe safeguards that have been put in place for responsible release of data or models that have a high risk for misuse (e.g., pre-trained language models, image generators, or scraped datasets)?
    \item[] Answer: \answerNA{} % Replace by \answerYes{}, \answerNo{}, or \answerNA{}.
    \item[] Justification: The paper does not introduce or release a new high-risk pretrained generative model, scraped dataset, or deployment system requiring additional release safeguards.
    \item[] Guidelines:
    \begin{itemize}
        \item The answer \answerNA{} means that the paper poses no such risks.
        \item Released models that have a high risk for misuse or dual-use should be released with necessary safeguards to allow for controlled use of the model, for example by requiring that users adhere to usage guidelines or restrictions to access the model or implementing safety filters. 
        \item Datasets that have been scraped from the Internet could pose safety risks. The authors should describe how they avoided releasing unsafe images.
        \item We recognize that providing effective safeguards is challenging, and many papers do not require this, but we encourage authors to take this into account and make a best faith effort.
    \end{itemize}

\item {\bf Licenses for existing assets}
    \item[] Question: Are the creators or original owners of assets (e.g., code, data, models), used in the paper, properly credited and are the license and terms of use explicitly mentioned and properly respected?
    \item[] Answer: \answerYes{} % Replace by \answerYes{}, \answerNo{}, or \answerNA{}.
    \item[] Justification: We cite the original papers for all pretrained models, datasets, and benchmarks, and use them in accordance with their respective licenses and terms of use.
    \item[] Guidelines:
    \begin{itemize}
        \item The answer \answerNA{} means that the paper does not use existing assets.
        \item The authors should cite the original paper that produced the code package or dataset.
        \item The authors should state which version of the asset is used and, if possible, include a URL.
        \item The name of the license (e.g., CC-BY 4.0) should be included for each asset.
        \item For scraped data from a particular source (e.g., website), the copyright and terms of service of that source should be provided.
        \item If assets are released, the license, copyright information, and terms of use in the package should be provided. For popular datasets, \url{paperswithcode.com/datasets} has curated licenses for some datasets. Their licensing guide can help determine the license of a dataset.
        \item For existing datasets that are re-packaged, both the original license and the license of the derived asset (if it has changed) should be provided.
        \item If this information is not available online, the authors are encouraged to reach out to the asset's creators.
    \end{itemize}

\item {\bf New assets}
    \item[] Question: Are new assets introduced in the paper well documented and is the documentation provided alongside the assets?
    \item[] Answer: \answerNA{} % Replace by \answerYes{}, \answerNo{}, or \answerNA{}.
    \item[] Justification: The paper does not currently release a new dataset, benchmark, or model checkpoint as a new public asset.
    \item[] Guidelines:
    \begin{itemize}
        \item The answer \answerNA{} means that the paper does not release new assets.
        \item Researchers should communicate the details of the dataset\slash code\slash model as part of their submissions via structured templates. This includes details about training, license, limitations, etc. 
        \item The paper should discuss whether and how consent was obtained from people whose asset is used.
        \item At submission time, remember to anonymize your assets (if applicable). You can either create an anonymized URL or include an anonymized zip file.
    \end{itemize}

\item {\bf Crowdsourcing and research with human subjects}
    \item[] Question: For crowdsourcing experiments and research with human subjects, does the paper include the full text of instructions given to participants and screenshots, if applicable, as well as details about compensation (if any)? 
    \item[] Answer: \answerNA{} % Replace by \answerYes{}, \answerNo{}, or \answerNA{}.
    \item[] Justification: The paper does not involve new crowdsourcing experiments or research with human subjects.
    \item[] Guidelines:
    \begin{itemize}
        \item The answer \answerNA{} means that the paper does not involve crowdsourcing nor research with human subjects.
        \item Including this information in the supplemental material is fine, but if the main contribution of the paper involves human subjects, then as much detail as possible should be included in the main paper. 
        \item According to the NeurIPS Code of Ethics, workers involved in data collection, curation, or other labor should be paid at least the minimum wage in the country of the data collector. 
    \end{itemize}

\item {\bf Institutional review board (IRB) approvals or equivalent for research with human subjects}
    \item[] Question: Does the paper describe potential risks incurred by study participants, whether such risks were disclosed to the subjects, and whether Institutional Review Board (IRB) approvals (or an equivalent approval/review based on the requirements of your country or institution) were obtained?
    \item[] Answer: \answerNA{} % Replace by \answerYes{}, \answerNo{}, or \answerNA{}.
    \item[] Justification: The paper does not involve human-subject studies, so IRB or equivalent approval is not applicable.
    \item[] Guidelines:
    \begin{itemize}
        \item The answer \answerNA{} means that the paper does not involve crowdsourcing nor research with human subjects.
        \item Depending on the country in which research is conducted, IRB approval (or equivalent) may be required for any human subjects research. If you obtained IRB approval, you should clearly state this in the paper. 
        \item We recognize that the procedures for this may vary significantly between institutions and locations, and we expect authors to adhere to the NeurIPS Code of Ethics and the guidelines for their institution. 
        \item For initial submissions, do not include any information that would break anonymity (if applicable), such as the institution conducting the review.
    \end{itemize}

\item {\bf Declaration of LLM usage}
    \item[] Question: Does the paper describe the usage of LLMs if it is an important, original, or non-standard component of the core methods in this research? Note that if the LLM is used only for writing, editing, or formatting purposes and does \emph{not} impact the core methodology, scientific rigor, or originality of the research, declaration is not required.
    %this research? 
    \item[] Answer: \answerNA{} % Replace by \answerYes{}, \answerNo{}, or \answerNA{}.
    \item[] Justification: The core method does not use LLMs as an important, original, or non-standard component. The pretrained vision-language backbones (Qwen3-VL-2B-Instruct and InternVL3.5-2B) are the standard models that \textsc{V-Lumen} augments and are described in the paper. No LLM is used to generate or annotate training data, and evaluation uses rule-based accuracy without LLM-based judging.
    \item[] Guidelines:
    \begin{itemize}
        \item The answer \answerNA{} means that the core method development in this research does not involve LLMs as any important, original, or non-standard components.
        \item Please refer to our LLM policy in the NeurIPS handbook for what should or should not be described.
    \end{itemize}

\end{enumerate}